\def\year{2022}\relax
\documentclass[letterpaper]{article} 
\usepackage{aaai22}  
\usepackage{times}  
\usepackage{helvet}  
\usepackage{courier}  
\usepackage[hyphens]{url}  
\usepackage{graphicx} 
\usepackage{natbib}  
\usepackage{caption} 
\DeclareCaptionStyle{ruled}{labelfont=normalfont,labelsep=colon,strut=off} 
\usepackage{algorithm}
\usepackage{algorithmic}

\usepackage{newfloat}
\usepackage{listings}
\floatstyle{ruled}
\newfloat{listing}{tb}{lst}{}
\floatname{listing}{Listing}
\usepackage{amsmath}
\DeclareMathOperator*{\argmax}{arg\,max}

\usepackage{amssymb}

\usepackage{xcolor}

\definecolor{umigray}{rgb}{0.93,0.93,0.93}

\usepackage{subfig}
\usepackage{booktabs}

\definecolor{codegray}{RGB}{238, 238, 255}

\title{NoiseGrad --- Enhancing Explanations \\ by Introducing Stochasticity to Model Weights}
\author {
    Kirill Bykov*\textsuperscript{,\rm 1, \rm 2 },
    Anna Hedström*\textsuperscript{,\rm 1, \rm 2 },
    Shinichi Nakajima\textsuperscript{\rm 1, \rm 3 },
    Marina M.-C. H\"ohne\textsuperscript{\rm 1, \rm 2 } \\
}
\affiliations {
    \textsuperscript{\rm 1} ML Group, TU Berlin, Germany \\
    \textsuperscript{\rm 2} Understandable Machine Intelligence Lab \\
    \textsuperscript{\rm 3} RIKEN AIP, Tokyo, Japan \\
    
    kirill.bykov@campus.tu-berlin.de, anna.hedstroem@tu-berlin.de, nakajima@tu-berlin.de, marina.hoehne@tu-berlin.de
}

\begin{document}
\maketitle

\begin{abstract}

Many efforts have been made for revealing the decision-making process of black-box learning machines such as deep neural networks, resulting in useful local and global explanation methods. For local explanation, stochasticity is known to help: a simple method, called \emph{SmoothGrad}, has improved the visual quality of gradient-based attribution by adding noise to the input space and averaging the explanations of the noisy inputs.  In this paper, we extend this idea and propose \emph{NoiseGrad} that enhances both local and global explanation methods. Specifically, NoiseGrad introduces stochasticity in the weight parameter space, such that the decision boundary is perturbed. NoiseGrad is expected to enhance the local explanation, similarly to SmoothGrad, due to the dual relationship between the input perturbation and the decision boundary perturbation.
We evaluate NoiseGrad and its fusion with SmoothGrad --- \emph{FusionGrad} ---  qualitatively and quantitatively with several evaluation criteria, and show that our novel approach significantly outperforms the baseline methods. Both NoiseGrad and FusionGrad are method-agnostic and as handy as SmoothGrad using a simple heuristic for the choice of the hyperparameter setting without the need of fine-tuning.
\end{abstract}

\section{Introduction}
The ubiquitous usage of Deep Neural Networks (DNNs), fueled by their ability to generalize and learn complex nonlinear functions, has presented both researchers and practitioners with the problem of non-interpretability and opaqueness of Machine Learning (ML) models. 
This lack of transparency, coupled with the widespread use of these highly complex models in practice, represents a risk and a major challenge for the responsible usage of artificial intelligence, especially in security-critical areas, e.g. the medical field.
In response to this, the field of eXplainable AI (XAI) has emerged intending to make the predictions of complex algorithms comprehensible for humans.

One possible dichotomy of post-hoc explanation methods can be carried out on the basis of whether these methods refer to the global or local properties of a learning machine.
The local level XAI aims to explain a model decision of an \emph{individual} input \cite{guidotti2018survey}, for which various methods, such as Layer-wise Relevance Propagation (LRP) \cite{bach2015pixel}, GradCAM \cite{Selvaraju_2019}, Occlusion \cite{zeiler2014visualizing}, MFI \cite{vidovic2016feature}, Integrated Gradient \cite{sundararajan2017axiomatic}, have proven effective in explaining DNNs. 
In contrast, global explanations aim to illustrate the decision process as a whole, without the connection to individual data samples. Recently, methods belonging to the Activation-Maximization \cite{erhan2009visualizing} family of methods have become widely popular, such as DeepDream \cite{mordvintsev2015inceptionism}, GAN-generated explanations \cite{nguyen2016synthesizing} and Feature Visualization \cite{olah2017feature}.

\begin{figure*}[h!]
\centering
\includegraphics[scale=0.33]{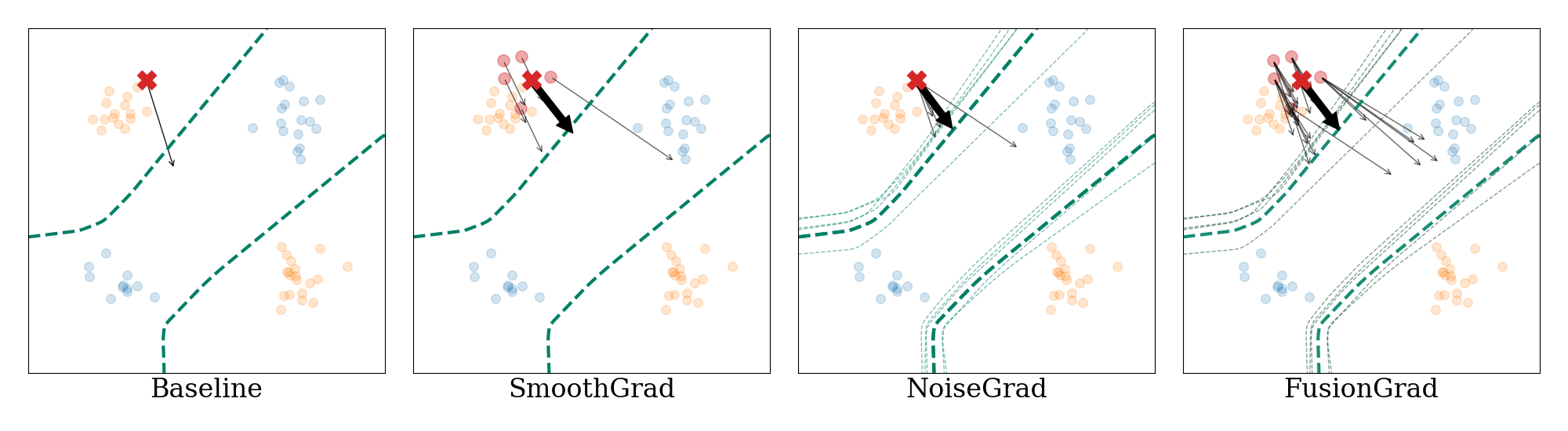}
\caption{Illustration of the differences in explanation behavior between Baseline (gradient-based explanation), SmoothGrad, NoiseGrad, and FusionGrad for a toy experiment.
Given 
training samples of two classes (orange and blue dots), a 3-layer MLP was trained for binary classification, where the learned decision boundary is shown by the green dashed line. The gradient explanations for a fixed test sample (red point) are shown as black arrows and the mean explanation as a bold black arrow. 
(a) For the baseline method, the explanation is the gradient itself. 
(b) SmoothGrad enhances the explanation by sampling points in the neighborhood (small red dots), and averaging their explanations (bold black arrow). (c) NoiseGrad enhances the explanation by averaging over perturbed models, indicated by multiple decision boundaries (thin green dashed lines). (d) FusionGrad combines SmoothGrad and NoiseGrad by incorporating both stochasticities in the input space and the model space.
}
\label{fig:overview_toy}
\end{figure*}

For local explanation, gradient-based methods are most popular due to their simplicity, however, they tend to suffer from the gradient shattering effect, which often results in noisy explanation maps \cite{samek2021explaining}. 
As a remedy, Smilkov et. al. proposed a simple method, called SmoothGrad \cite{smilkov2017smoothgrad}, where stochasticity is introduced to the input.  Specifically, it adds Gaussian Noise to the input features $n$ times, computes the $n$ corresponding explanations, and takes the average over the $n$ explanations.
SmoothGrad is applicable to any local explanation method and has been practically proven to reduce the visual noise in the explanation map.

The mechanism behind SmoothGrad's enhancement of explanations is not yet well understood. One could argue that SmoothGrad averages out the shattering effect. However, SmoothGrad performs best when the added noise level is around 10\%--20\% of the signal level, which not only smooths out peaky derivatives but is large enough to cross the decision boundary. From this fact, we hypothesize
that SmoothGrad perturbs the test sample in order to get a signal from the steepest part of the decision boundary. 
This motivated us to explore another way of using stochasticity: instead of adding noise to the input, our proposed method --- NoiseGrad (NG) --- draws samples from the network weights from a \emph{tempered} Bayes posterior \cite{Wenzel20}, such that the decision boundaries of some models are close to the test sample, which results in more precise explanations.

Our hypothesis leads to a natural and easy way of hyperparameter choice: the noise level added to the weights (which corresponds to the temperature of the tempered Bayes posterior) is chosen such that the relative performance drop is around $5\%$. 
In addition, we approximate the tempered Bayes posterior by multiplicative noise applied to the network weights --- in the same spirit as MC dropout \cite{Gal16}.
Thus, our proposed method NoiseGrad can be implemented as easily as SmoothGrad with an automatic hyperparameter choice and is applicable to any model architecture and explanation method.
Our experiments empirically support our hypothesis and show quantitatively and qualitatively that NoiseGrad outperforms SmoothGrad and combining NoiseGrad with SmoothGrad, which we refer to as FusionGrad, further boosts the performance.
An overview of our proposed methods is shown on a toy experiment in Figure \ref{fig:overview_toy}.

Another advantage of NoiseGrad over SmoothGrad is that it is straightforwardly applicable to global explanations as well.  For example, we can replace the objective function for activation maximization with its average over the model samples, which is expected to stabilize the image representing the features captured by neurons. Our experiments demonstrate that NoiseGrad improves global explanations in terms of human interpretability and vividness of illustrated abstractions.

Our main contributions include:
\begin{itemize}
    \item We propose a novel method, \emph{NoiseGrad}, that improves local and global explanation methods by introducing stochasticity to the model parameters.
    \item The performance gain by NoiseGrad and its fusion with SmoothGrad, \emph{FusionGrad}, for local explanations, is shown qualitatively and quantitatively using different evaluation criteria.
    \item We observe that NoiseGrad is further capable of enhancing global explanation methods.
\end{itemize}

\section{Background} 

Let $f(\cdot;\hat{W}): \mathbb{R}^d \rightarrow \mathbb{R}^k$ be a 
neural network with learned weights $\hat{W} \subset \mathbb{R}^S$ that maps a vector $\boldsymbol{x} \in \mathbb{R}^d$ from the input domain to a vector $y \in \mathbb{R}^k$ in the output domain. In general, attribution methods could be viewed as an operator $E\left(\boldsymbol{x}, f(\cdot, \hat{W})\right)$ that attributes relevances to the features of the input $\boldsymbol{x}$ with respect to the model function $ f(\cdot, W)$. More in-depth discussion about the different explanation methods used can be found in the Appendix.

\paragraph{Enhancing local explanations by adding noise to the inputs}

A recently proposed popular method, called
SmoothGrad (SG), seeks to alleviate noise and visual diffusion of saliency maps by introducing stochasticity to the inputs \cite{smilkov2017smoothgrad}. 
SmoothGrad adds Gaussian noise to the input and takes the average over $N$ instances of noise:
\begin{equation}
\begin{split}
E_{SG}\left(\boldsymbol{x}\right) = \textstyle \frac{1}{N} \sum_{i = 1}^{N}E\left(\boldsymbol{x} + \xi_i, f(\cdot, \hat{W})\right), \\ \xi_i \sim \mathcal{N}(\mathbf{0},\sigma_{\mathrm{SG}}^{2} \mathbf{I})
\end{split}
\end{equation}
where $\mathcal{N}(\mathbf{\mu},\mathbf{\Sigma})$ is the Normal distribution
with mean $\mathbf{\mu}$ and covariance $\mathbf{\Sigma}$ and $\mathbf{I}$ the identity matrix.
The authors of the original paper state that SG allows smoothening the gradient landscape, thus providing better explanations. 
Later SmoothGrad has also proven to be more robust against adversarial attacks \cite{dombrowski2019explanations}. 

\paragraph{Enhancing explanations by approximate Bayesian learning}

From a statistical perspective, training DNNs with the most commonly used loss functions and regularizers, such as categorical cross-entropy for classification and MSE for regression, can be seen as performing \emph{maximum a-posteriori (MAP) learning}. Hence, the resulting weights can be thought of as a point estimate for a posterior mode in the parameter space, capturing no uncertainty information. 
Recently, \citet{bykov2021explaining} showed that incorporating information about posterior distribution can enhance local explanations for DNNs. Intuitively, in contrast to the MAP learning, where point estimates of weights represent one deterministic decision-making strategy, a posterior distribution represents an infinite ensemble of models, which employ different strategies towards the prediction. By aggregating the variability of the decision-making processes of networks, we can obtain a broader outlook on the features that were used for the prediction, and thus deeper insights into the models' behavior \cite{grinwald2022visualizing}.

Since exact Bayesian Learning is intractable for DNNs, a plethora of approximation methods have been proposed, e.g., Laplace Approximation \cite{ritter2018scalable}, Variational Inference \cite{Graves11,Osawa19}, MC dropout \cite{Gal16}, Variational Dropout \cite{Kingma15,Molchanov17}, MCMC sampling \cite{Wenzel20}.
Since most of the approximation methods require full retraining of the network or evaluation of the second-order statistics, which are computationally expensive, we use a cruder approximation with multiplicative noise to draw model samples for NoiseGrad.

\section{Method}

As mentioned previously, the mechanism why SmoothGrad improves explanations has not been well understood. 
In empirical experiments we found that SmoothGrad with a recommended 10\%--20\% noise level is large enough to cross the decision boundary, resulting in a significant classification accuracy drop.
This finding implies that SmoothGrad does not only smooth the peaky derivative but also collects signals from the steepest part of the likelihood, i.e., decision boundary, by perturbing the input sample with large noise.

Motivated by this observation, we propose another way of introducing stochasticity -- instead of perturbing the input, we perturb the model itself. More specifically, we propose a new method --- NoiseGrad --- which draws network weight samples from a \emph{tempered} Bayes posterior \cite{Wenzel20}, i.e.,
the Bayes posterior with a temperature higher than 1.
The temperature should be so high that the decision boundaries of some model samples are close to the test sample, which reinforces the signals for explanations. 

\begin{figure}[t!]
\centering
\includegraphics[width=\linewidth]{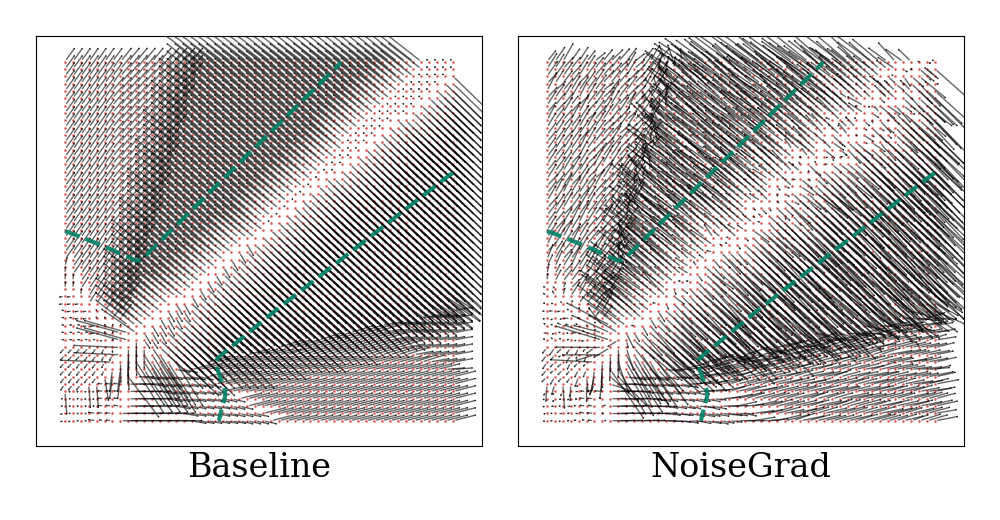}
\vspace{-0.7cm}
\caption{Illustration of the impact of NoiseGrad on a gradient flow map. For the same problem as in Figure \ref{fig:overview_toy}, for each grid-point the gradient is computed (left). On the right we can observe the effect of NoiseGrad --- it smoothens the gradients by perturbing the decision boundary.}
\label{fig:ng_smoothing}
\end{figure}

\paragraph{Local explanation with NoiseGrad}
Mathematically, we define the local explanation with NoiseGrad (NG) as follows
\begin{equation}
    E_\mathrm{NG}(\boldsymbol{x}) =
    \textstyle
    \frac{1}{M} \sum_{i = 1}^{M} E\left(\boldsymbol{x}, f(\cdot, \mathcal{W}_i)\right),
 \end{equation}
where $\{\mathcal{W}_i\}, i \in [1, M]$ are samples drawn from a tempered Bayes posterior.
Since approximate Bayesian learning is computationally expensive, we approximate the posterior with multiplicative Gaussian noise -- in the spirit of MC dropout \cite{Gal16}:
 $\mathcal{W}_i = \hat{W} \cdot \eta_i,$ with $\eta_i\sim  \mathcal{N}(\mathbf{1},\,\sigma_{\mathrm{NG}}^{2}\mathbf{I})$, where $\mathbf{I}$ refers to an identity matrix.
By averaging over a sufficiently large number of samples $M$, we expect NG to smooth the signal and also to collect amplified signals from models whose decision boundary is close to the test sample. This and the smoothing capabilities of the NG method can be observed in Figure \ref{fig:ng_smoothing}, where the gradients are plotted for each grid-point on the toy dataset used before in Figure \ref{fig:overview_toy} for both Baseline and NG. From the results shown, we can observe that NoiseGrad in fact smoothens out the gradient.

\paragraph{FusionGrad}
We also propose FusionGrad (FG), a combination of NoiseGrad and SmoothGrad, to incorporate both stochasticities in the input space and the model space
\begin{equation} \label{eq:ngplus}
    E_\mathrm{FG}(\boldsymbol{x}) =
    \textstyle
    \frac{1}{N} \sum_{i = 1}^{N} \frac{1}{M} \sum_{j = 1}^{M} E\left(\boldsymbol{x} + \xi_j, f(\cdot, \mathcal{W}_i)\right)
\end{equation}
where $\xi_j \sim \mathcal{N}(\mathbf{0},\,\sigma_{\mathrm{SG}}^{2} \mathbf{I})$, $N$ the number of noisy inputs, and $M$ the number of model samples.
We show in our experiments that FG further boosts the performance of NG, providing the best qualitative and quantitative performances.

\paragraph{Global explanation with NoiseGrad}

Unlike SmoothGrad, NoiseGrad  
can be used to enhance global explanation methods.
An important class of global explanation methods is activation maximization (AM) \citep{erhan2009visualizing},
which synthetically creates an input that maximizes a given function $g(x)$.
Usually, the activation of a particular neuron is maximized, and thus the generated input can embody the main concepts and abstractions that the DNN is looking for. 
Many variants with different types of regularization have emerged \cite{nguyen2016synthesizing, olah2017feature}
---
regularization is necessary because otherwise, AM might synthesize adversarial inputs, which would not convey visual information to the investigator.

NoiseGrad enhancement of any AM technique could be performed as follows: given the target function $g(x, \hat{W})$ for a model $f(\cdot;\hat{W})$, we 
sample $M$ models with the NG procedure, and maximize the average function over the number of perturbed models:

\begin{equation} \label{eq:ng_global_opt}
\textstyle
    \argmax_{x \in \mathcal{C}} \frac{1}{M}\sum_{i = 1}^M g(x, \mathcal{W}_i),
\end{equation}
where $ \mathcal{C}$ is a regularized input domain, specific to a particular implementation of an AM method.

\paragraph{Heuristic for hyperparameter setting}

One of the reasons for the popularity of SmoothGrad
is that it does not require hyperparameter tuning:
it works well if the number $N$ of noisy samples is sufficiently large, and the input noise level $\sigma_{\mathrm{SG}}$ is set to a value in the recommended range, 10\%--20\%, compared to the signal level.

A major question is if one can set the noise level 
$\sigma_{\mathrm{NG}}$ for NoiseGrad in a similar way that does not require fine-tuning.
We put forward a simple hypothesis: since we need signals from models whose decision boundaries are close to the test sample, we might choose the noise level $\sigma_{\mathrm{NG}}$ such that we observe a certain performance drop. For the classification setting with balanced classes, we can use accuracy as a performance measure: from experimental results (discussed more in-depth in the Appendix) we recommend to set the relative accuracy drop $\textrm{AD} (\sigma) = 1 - (\textrm{ACC}(\sigma) - \textrm{ACC}(\infty))/(\textrm{ACC}(0) - \textrm{ACC} (\infty))$ to around $5\%$,
where $\textrm{ACC}(\sigma)$ denotes the classification accuracy at the noise level $\sigma$. Note that $\textrm{ACC}(0)$ and $\textrm{ACC}(\infty)$ correspond to the original accuracy and the chance level, respectively. 
This rule of thumb can be used for various model architectures with different scales,
as shown in the next section.

As a heuristic for FusionGrad, we recommend  to  half  both $\sigma_{\mathrm{SG}}$
and $\sigma_{\mathrm{NG}}$
(as found by their respective heuristics) to equal the contribution from the input perturbation and the model perturbation.
Further, we empirically found that $10$ samples are sufficient for both methods.
With those heuristics, NoiseGrad and FusionGrad can be used as effortlessly as SmoothGrad.
A detailed discussion on the relationship between the explanation quality (localization criteria) and accuracy drop can be found in the Appendix.
 
\begin{table*}[h!]
  \centering
  \begin{tabular}{lcccc}
    \toprule
    {Method}     & {Localization ($\uparrow$)}    & {Faithfulness ($\uparrow$)}  &  {Robustness ($\downarrow$)} & {Sparseness ($\uparrow$)}  \\
    \midrule
  
Baseline &	  0.7315 $\pm$ 0.0505 & 0.3413 $\pm$ 0.1549 & 0.0763 $\pm$  0.0265 & \textbf{0.6272 $\pm$ 0.0475} \\
SG &	  0.8263 $\pm$ 0.0483 & 0.3465 $\pm$ 0.1601 & 0.0590  $\pm$ 0.0235 & 0.5310 $\pm$ 0.0635 \\
NG 	&  0.8349 $\pm$ 0.0367 & \textbf{0.3635 $\pm$ 0.1536} &0.0224  $\pm$ 0.0080 & 0.5794 $\pm$ 0.0533 \\
FG &  \textbf{0.8435} $\pm$ \textbf{0.0358} & \textbf{0.3697 $\pm$ 0.1465} & \textbf{0.0153  $\pm$ 0.0058 }& 0.5721 $\pm$ 0.0532 \\
  \end{tabular}
  \caption{Comparison of attribution quality where the noise levels are set by the heuristic.  $\uparrow$ and $\downarrow$ indicate the larger is the better and the smaller is the better, respectively. 
The values of the best method and the methods that are not significantly outperformed by the best method,
according to the Wilcoxon signed-rank test for $p=0.05$,
are bold-faced.
  }
  \label{sample-table}
\end{table*}

\begin{table*}[h!]
  \centering
  \begin{tabular}{lcccccc}
    \toprule
    Method & LeNet  & VGG11 & VGG16 & RN9 & RN18 &  RN50  \\
    \midrule
    Baseline & 0.922 $\pm$ 0.033 & 0.961 $\pm$ 0.017 & 0.967 $\pm$ 0.015 & 0.926 $\pm$ 0.026 & 0.913 $\pm$ 0.04 & 0.912 $\pm$ 0.035  \\
    SG & 0.940 $\pm$ 0.048 & 0.971 $\pm$ 0.025 &   0.963 $\pm$ 0.034 & 0.975 $\pm$ 0.015 & 0.962 $\pm$ 0.026 & 0.967 $\pm$ 0.022 \\
    NG & 0.949 $\pm$ 0.029 & 0.982 $\pm$ 0.011 &  \textbf{0.985} $\pm$ \textbf{0.011} & 0.978 $\pm$ 0.012 & 0.963 $\pm$ 0.023 & \textbf{0.973} $\pm$ \textbf{0.017}  \\
    FG & \textbf{0.961} $\pm$ \textbf{0.025} & \textbf{0.984} $\pm$ \textbf{0.011}  & 0.982 $\pm$ 0.013 & \textbf{0.982} $\pm$ \textbf{0.011} & \textbf{0.975} $\pm$ \textbf{0.016} & 0.969 $\pm$ 0.019 \\
    \bottomrule
  \end{tabular}
  \caption{Attribution ranking scores ($\mathrm{AUC}$) for different architectures with noise levels set by our proposed heuristic. We can observe that either NG or FG outperforms Baseline and SG. For the sake of space, we refer ResNets as RNs.}
  \label{table-heuristic}
\end{table*}

\section{Experiments On Local Explanations}
\label{sec:ExperimentsLocalExplanation}

In this section, we explain datasets and evaluation metrics used for evaluating our proposed methods for
local \textit{attribution quality}.

\paragraph{Datasets}

To measure the goodness of an explanation, one typically needs to resort to proxies for evaluation since no ground-truth for explanations exists. Similar to \citet{arras2021ground} and \citet{BAM2019, github_MNIST_segmentation}, we therefore design a controlled setting for which the ground-truth segmentation labels are simulated. For this purpose, we construct a semi-natural dataset CMNIST (customized-MNIST), where each MNIST digit \cite{lecun2010mnist} is displayed on a randomly selected CIFAR background \cite{krizhevsky2009learning}. To ensure that the explainable evidence for a class lies in the vicinity of the object itself, rather than in its contextual surrounding, we uniformly distribute CIFAR backgrounds for each MNIST digit class as we construct the CMNIST dataset. Ground-truth segmentation labels for the explanations are formed by creating different variations of segmentation masks around the object of interest such as a squared box around the object or the pixels of the object itself. Moreover, to understand the real impact of SOTA, we use the PASCAL VOC 2012 object recognition dataset \cite{Everingham10} and ILSVRC-15 dataset \cite{ILSVRC15} for evaluation, where object segmentation masks in the forms of bounding boxes are available. Further details on training- and test splits, preprocessing steps and other relevant dataset statistics can be found in the Appendix. 

In an explainability context, the question naturally arises whether object localization masks can be used as ground-truth labels for explanations of natural datasets in which the independence of the models from the background cannot be guaranteed. We, therefore, report quantitative metrics only on the controlled semi-natural dataset but report qualitative results on the natural dataset as well. 

\paragraph{Evaluation metrics}

While the debate of what properties an attribution-based explanation ought to fulfill continues, several works \cite{montavon2018methods,alvarez2018towards,carvalho2019machine}
suggest that in order to produce human-meaningful explanations one metric alone is not sufficient. To broaden the view of what it means to provide a good explanation, we evaluate the explanation-enhancing methods using four well-studied properties --- \textit{localization} \cite{zhang2016topdown, kohlbrenner2020best, theiner2021interpretable, arras2021ground}, \textit{faithfulness} \cite{bach2015pixel, samek2016evaluating, bhatt2020evaluating, nguyen2020quantitative, rieger2020simple}, \textit{robustness} \cite{alvarez2018towards, montavon2018methods, yeh2019fidelity}, and \textit{sparseness} \cite{nguyen2020quantitative, chalasani2020concise, bhatt2020evaluating}. While there exists several empirical interpretations, or operationalizations, for each of these qualities, we selected one metric per category. We adopted \textit{Relevance Rank Accuracy} \cite{arras2021ground} to express localization of the attributions, \textit{Faithfulness correlation} \cite{bhatt2020evaluating} to capture attribution faithfulness, applied \textit{max-Sensitivity} \cite{yeh2019fidelity} to express attribution robustness and \textit{Gini index} \cite{chalasani2020concise} to assess the sparsity of the attributions. All evaluation measures are clearly motivated, defined, and discussed in the Appendix.

\paragraph{Explanation methods}

NoiseGrad is \textit{method-agnostic}, which means, that it can be applied in conjunction with \textit{any} explanation method. However, in these experiments, we focus on a popular category of post-hoc gradient-based attribution methods and use \textit{Saliency} (SA) \cite{morch} as the base explanation method in the experiments similar as in \cite{smilkov2017smoothgrad}. Since the majority of model-aware local explanation methods make use of the model gradients, we argue that a potential explanation improvement on SA with our proposed method may also be transferred to an improvement of a related gradient-based explanation method. As the comparative baseline explanation method (Baseline), we employ the Saliency explanation, which adds no noise to either the weights nor the input. We report results on additional explanation methods in the Appendix.

\paragraph{Model architectures}

Explanations were produced for networks of different architectural compositions such as ResNet \cite{he2015deep}, VGG \cite{simonyan2014very} and LeNet \cite{lecun1998gradient}. All networks were trained for image classification tasks so that they showcased a comparable test accuracy to a minimum of $86$\%, $92$\% and $86$\% classification accuracy for CMNIST, PASCAL VOC 2012, and ILSVRC-15 datasets respectively.  For more details on the model architectures, optimization configurations, and training results, we refer to the Appendix.

\subsection{Results}
In the following, we present our experimental results. The findings can be summarized as follows: (i) both NG and FG offer an advantage over SG measured with several metrics of attribution quality and (ii) as a heuristic, choosing the hyperparameters for NG and FG according to a classification performance drop of 5\% typically result in explanations with a high attribution quality.

\begin{figure*}[h!]
\centering
\includegraphics[width=0.78\linewidth]{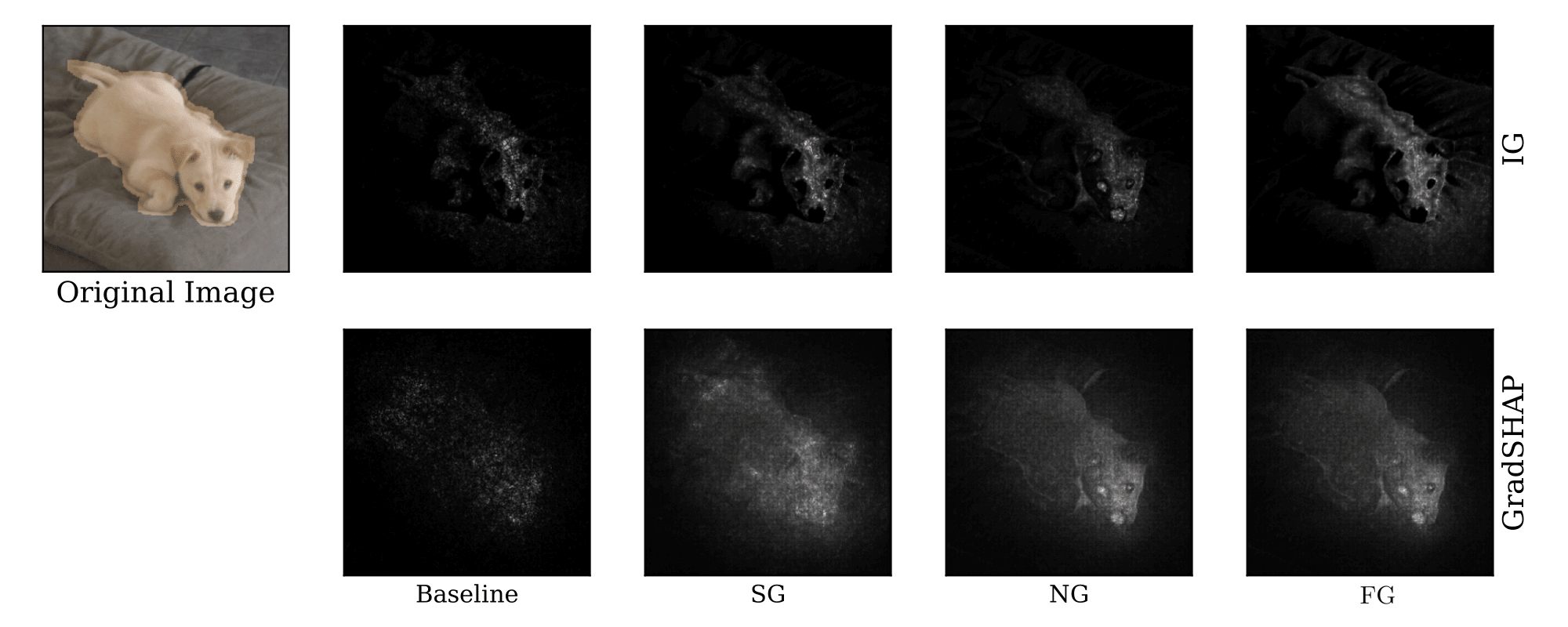}
\caption{Attribution maps by Baseline, SG, NG, and FG for two base explanation methods
Integrated Gradients (IG) and GradientSHAP (GradSHAP), 
for an image from the PASCAL VOC 2012 dataset. We can observe that both NG and FG improve the sharpness of the attributions compared to Baseline and SG. Moreover, NG highlights semantic features of the dog, such as the nose and the eyes, which are not visible for Baseline and SG.
}
\label{fig:XAIMethodComparison}
\end{figure*}
\begin{figure*}[h!]
\centering
\includegraphics[width=0.78\linewidth]{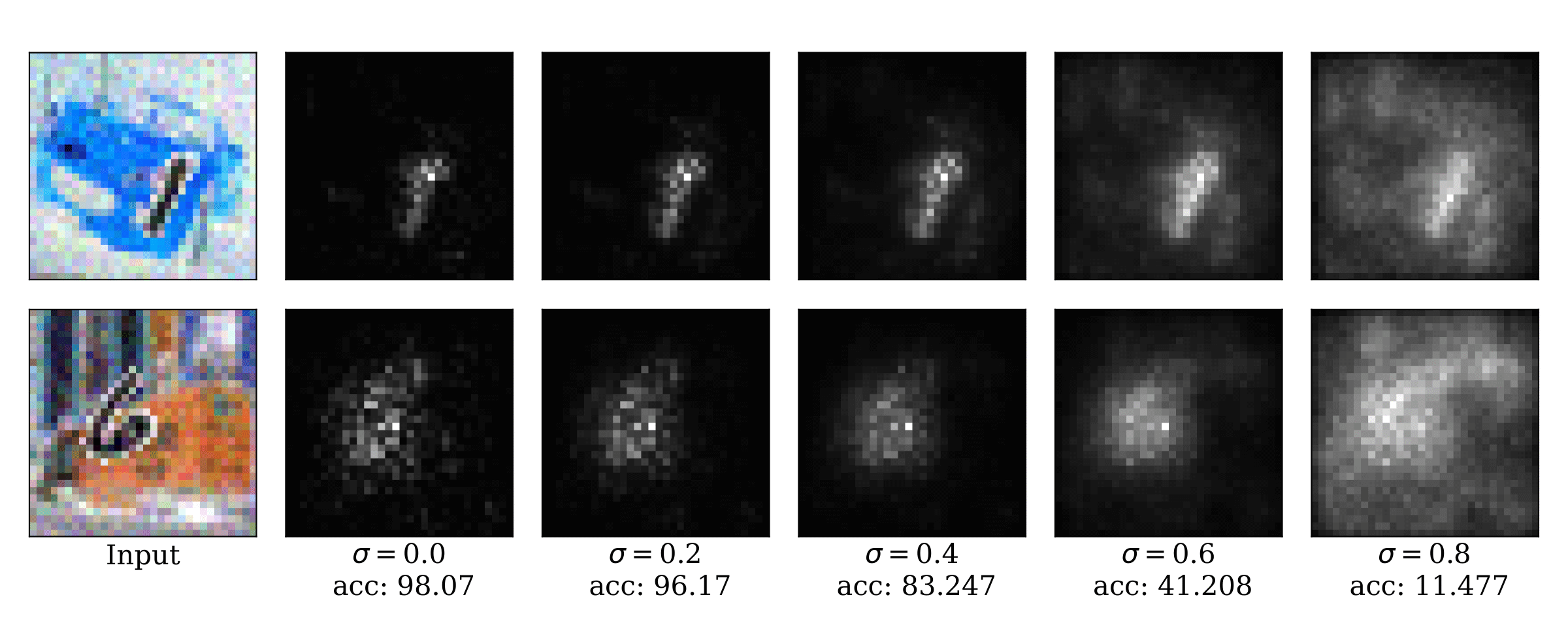}
\caption{
Illustration of NG-enhanced Saliency explanations for the CMNIST dataset: 
we observe an improvement of the localization ability of the explanation when increasing the hyperparameter $\sigma$ until $\sigma\leq0.4$ –– afterward, if the noise amplitude becomes too large, the models lose their predicting ability, which results in noisy attribution maps.
}
\label{fig:CMNIST}
\end{figure*}

\paragraph{Quantitative evaluation} 
We start by examining the performance of the methods considering the four aforementioned attribution quality criteria applied to the absolute values of their respective explanations. The results are summarized in Table \ref{sample-table}, where the methods (Baseline, SG, NG, FG) are stated in the first column and the respective values for localization, faithfulness robustness, and sparseness in columns 2-5.
The scores were computed and averaged over 256 randomly chosen test samples from the CMNIST dataset, using a ResNet-9 classifier and the Saliency as the base attribution method. The Quantus library was employed for XAI evaluation\footnote{Code can be found at \url{https://github.com/understandable-machine-intelligence-lab/quantus}} \cite{hedstrom2022quantus}. 

The noise level for SG, NG, and FG is set by the heuristic, which was described in the method section.
We conducted the same experiment with additional base attribution methods and datasets and found similar tendencies, which are reported in the Appendix. In the Appendix, we also investigated how the different noise levels for FusionGrad ($\sigma_{\mathrm{NG}}$ and $\sigma_{\mathrm{SG}}$) influence the ranking of attributions, as well as performed model parameter randomization sanity checks \cite{adebayo2018sanity}.

From Table \ref{sample-table}, we can observe a significant attribution quality boost by our proposed methods, NG, and FG in comparison to the baselines, Baseline and SG. For each of the examined quality criteria, the values range between $[0, 1]$. For localization, faithfulness, and sparseness higher values are better and for robustness lower values are better. The combination of SmooothGrad and NoiseGrad, i.e., FusionGrad is \emph{significantly} better than either method alone. 
In summary, we conclude that NG outperforms SG on all four criteria and Baseline on the most criteria except Sparseness,
and FG further boosts the performance.
In general, any perturbation naturally degrades the sparseness,
and therefore Baseline gives the best sparseness score.  
Note that NG and FG both improve the other criteria with less degradation of sparseness in comparison to SG.
It is important to also emphasize that evaluation of explanation methods should always be viewed holistically --- i.e., while Baseline may be the most sparse explanation, it would \textit{not} be the overall preferred option since it is the least faithful, localized, and robust explanation of them all. Further evaluation results on the ILSVRC-15 dataset can be found in the Appendix.

\begin{figure}[t]
\centering
\includegraphics[scale=0.28]{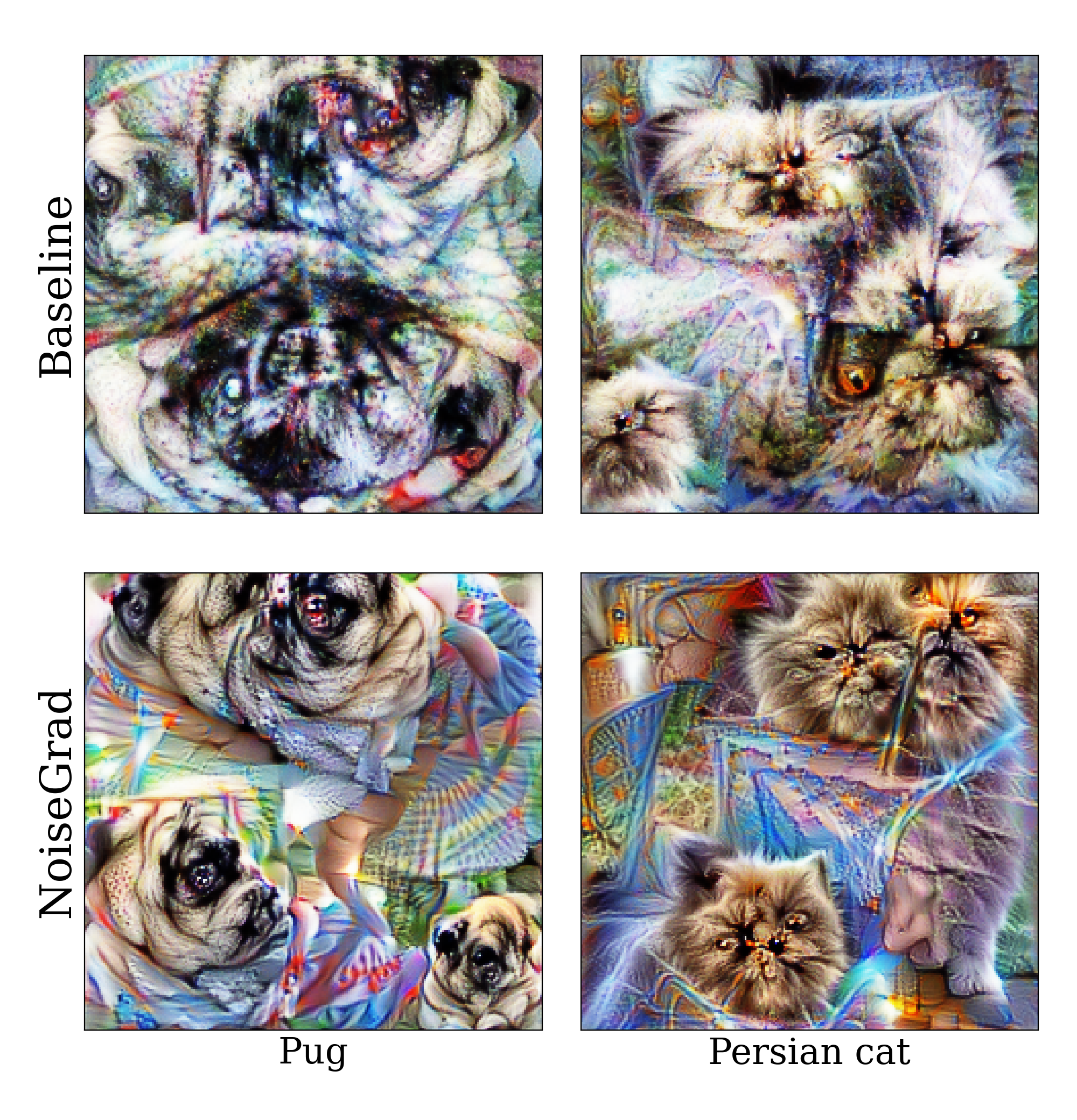} 
\caption{Global explanation by activation maximization (AM) without (top row) and with (bottom) NoiseGrad, applied to ResNet-18 network pre-trained on ImageNet dataset. 
For each column, the output neuron for the specified class is explained.
}
\label{fig:noisegrad_global}
\end{figure}

\paragraph{Heuristic applied to different architectures} 
In Table \ref{table-heuristic}, we present ranking $\mathrm{AUC}$ scores for different model architectures trained on CMNIST, using the recommended heuristic to set the noise level. 

We can observe that appropriate noise levels are chosen with the proposed heuristic --- where NG and FG significantly outperform Baseline and SG.

\paragraph{Qualitative evaluation} 
Figure \ref{fig:XAIMethodComparison}
shows attribution maps for an image from the PASCAL VOC 2012 dataset for Baseline, SG, NG, and FG for two attribution methods, 
Integrated Gradients (IG) \cite{sundararajan2017axiomatic} and GradientSHAP (GradSHAP) \cite{lundberg2017unified}.
Compared to Baseline and SmoothGrad, NG and FG demonstrate improved localized attribution with improved vividness. Semantic meaningful features such as the nose and the eyes of the dog are highlighted by NG but not by SG, indicating that our methods can find additional attributional evidence for a class that SG or Baseline explanation does not. Furthermore, as we enumerated several test samples to find representative qualitative characteristics that distinguish the different approaches, we could conclude that attributions of NG and FG are typically more crisp and concise compared to Baseline and SG explanations.

Figure \ref{fig:CMNIST} shows the noise level dependence of the NG attribution map with Saliency as the base attribution. Visually, the attribution seems to improve with a noise level between $\sigma_{\mathrm{NG}} \in [0.2, 0.4]$, which is chosen by our heuristic as well.  More examples are given in the Appendix.

\section{Experiments On Global Explanations}

Finally, we apply NoiseGrad to enhance the global explanations generated by Feature Visualisation \footnote{For generating global explanations following library was used https://github.com/Mayukhdeb/torch-dreams} \cite{olah2017feature}.
We applied FV to the output neurons for different classes
of a ResNet-18 network pre-trained on ImageNet dataset with and without NoiseGrad.

Figure \ref{fig:noisegrad_global} indicates the feature visualization images by 
the Baseline AM (top row) and by AM using NoiseGrad (bottom row). We can observe that the visualized abstractions with NoiseGrad are more vivid and more human-understandable, implying that the NG-enhanced global explanation can convey improved recognizability of underlying high-level concepts.
More examples and experiments can be found in the Appendix.

\section{Conclusion}\label{sec:Discussion}
In this work, we demonstrated that the use of stochasticity in the parameter space of deep neural networks can enhance techniques for eXplainable AI (XAI). 

Our proposed NoiseGrad draws samples from the approximated tempered Bayes posterior, such that the decision boundary of some model samples is close to the test sample, effectively amplifying the gradient signals.  
In our experiments on local explanations, we have shown the advantages of NoiseGrad and its fusion with the existing SmoothGrad method qualitatively and quantitatively on several evaluation criteria. 
A notable advantage of NoiseGrad over SmoothGrad is that it can also enhance global explanation methods by smoothing the objective for activation maximization (AM), leading to enhanced human-interpretable concepts learned by the model.
We believe that our idea of introducing stochasticity in the parameter space facilitates the development of practical and reliable XAI for real-world applications.

\paragraph{Limitations} 
Since the number of parameters in a DNN is usually larger than the number of features in the input data, NoiseGrad is more computationally expensive than SmoothGrad (more discussion in the Appendix). 

In addition, explanation evaluation is still an unsolved problem in XAI research and each evaluation technique comes with individual drawbacks. Further research is needed to establish a sufficient set of quantitative evaluation metrics beyond the four criteria used in this paper.
\paragraph{Future work} To broaden the applicability of our proposed methods, we are interested to investigate the performance of NG and FG on other tasks than image classification such as time-series prediction or NLP. 
We also want to further explore how NG and FG explanations change when alternative ways of adding noise to the weights of a neural network are employed e.g., by adding different levels of noise to different layers or individual neurons.

\section*{Acknowledgements}
This work was partly funded by the German Ministry for Education and Research through the third-party funding project Explaining 4.0 (ref. 01IS20055) and as BIFOLD
- Berlin Institute for the Foundations of Learning and Data 
(ref. 01IS18025A and ref 01IS18037A).


\bibliography{refs.bib}
\newpage
\newpage

\appendix

\section{Appendix}

\begin{figure*}[h!]
\centering
\includegraphics[scale=0.3]{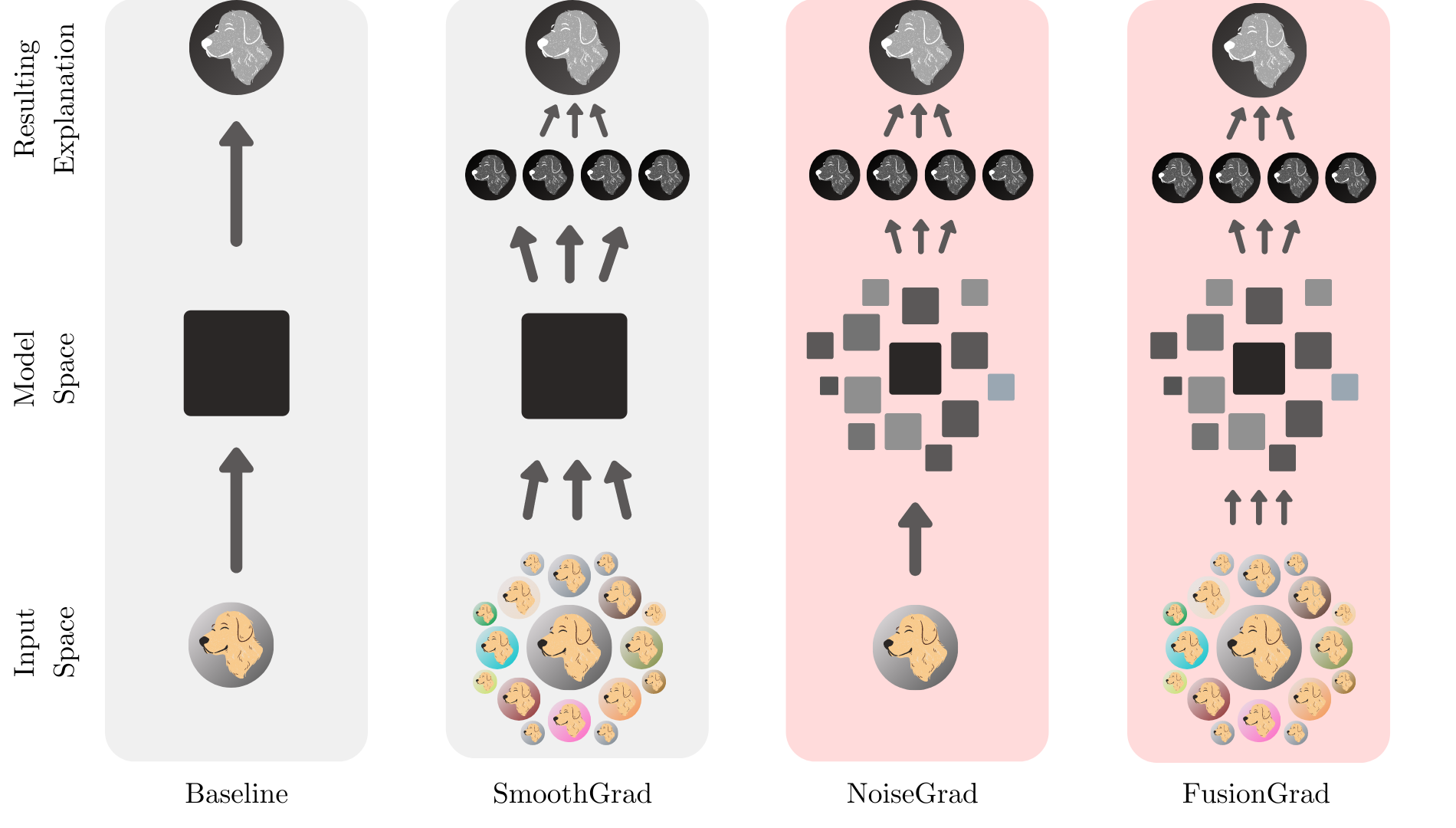}
\caption{Illustration of our proposed methods: the functionality of the individual methods, i.e., Baseline, SmoothGrad, NoiseGrad, and FusionGrad are visualized schematically from left to right, each partitioned in input space, model space, and the resulting explanation from bottom to top. The baseline explanations are computed in a deterministic fashion -- one input (dog), one model (black square), one explanation. SmoothGrad enhances the explanation by exploring the neighborhood of a datapoint, here indicated by multiple noisy versions of the input. In contrast, our proposed method, NoiseGrad, enhances the explanations by investigating the neighborhood of the trained model, indicated by multiple versions of the model. FusionGrad combines SmoothGrad and NoiseGrad by incorporating both stochasticities in the input space and model space.}
\label{fig:overview}
\end{figure*}

\subsection{Toy experiment}
For the toy experiment, the dataset was generated, consisting of a total of 1024 data points, generated from an equal mixture of 4 Gaussian distributions. All 4 distributions have diagonal covariance matrix, with diagonal elements equal to 0.5 and mean values of $[8,8], [1, 8], [8, 1], [1, 1],$ where the first 2 distributions were assigned a label 0 and last 2 distributions assigned with label 1.

Dataset was randomly split into test and training parts, test dataset consisting of 64 data points. 3-layer MLP network with ReLU activation was trained using Stochastic Gradient Descend algorithm and achieves 100 percent accuracy.

\begin{figure*}[h!]
\centering
\includegraphics[scale=0.3]{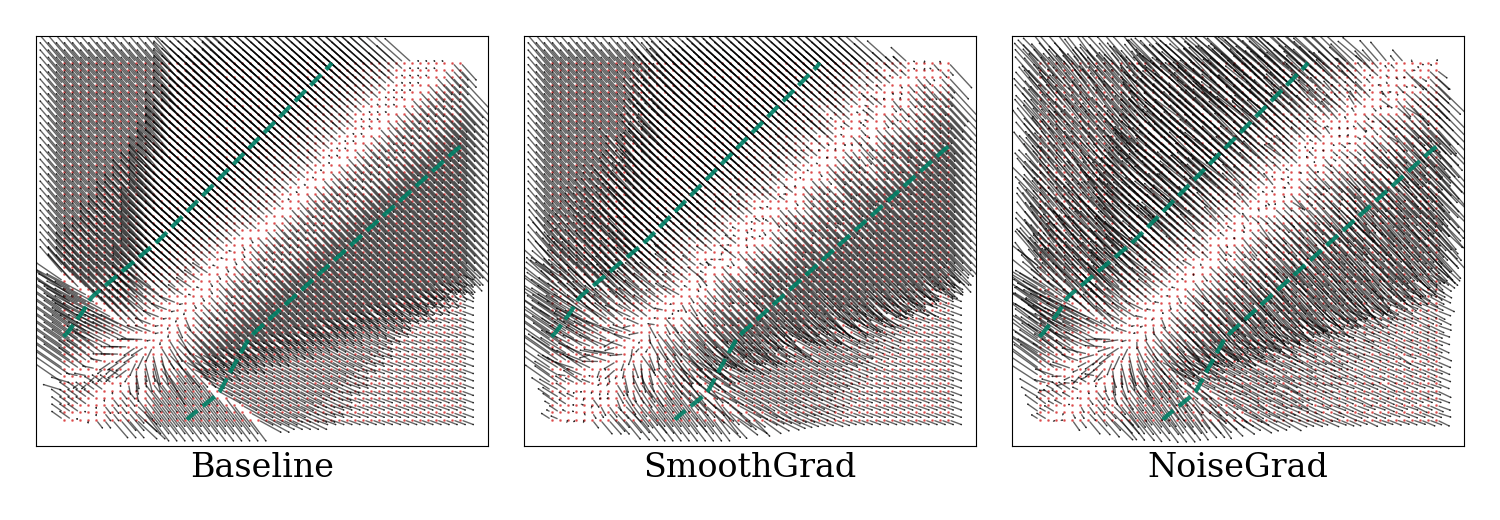}
\caption{Comparison of smoothing effect on gradient between SmoothGrad and NoiseGrad. From left to right: standard gradient flow, SmoothGrad enhanced gradient map and NoiseGrad enhanced gradients.}
\label{fig:SGvsNG}
\end{figure*}

As described in the main paper, we have observed that NoiseGrad smoothens the gradients. In comparison with a SmoothGrad method, that intrinsically smoothens the gradient with convolution operation in the neighborhood of the original data point, NoiseGrad perturbs decision bound. This, in turn, results in interesting observation that could be illustrated in Figure \ref{fig:SGvsNG} where we can observe that SG has constant gradients in the areas, where all standard (baseline) gradients have the same value, while in NG this is not happening, and all gradient in these areas slightly differ.

\subsection{Heuristic for choosing hyperparameters}
\label{sec:heuristic}

For each explanation-enhancing method, we clarify the recommended heuristic i.e, the rule of thumb that was applied in the experiments to choose the right level of noise.

\paragraph{SmoothGrad}
As suggested by the authors in \cite{smilkov2017smoothgrad}, we set the standard deviation of the input noise as follows
\begin{align}
\sigma_{\mathrm{SG}} = \alpha_{\mathrm{SG}}(\max(x) - \min(x)),
\label{eq:NoiseLevelSG}
\end{align}
where $x$ is the input image and $\alpha_{\mathrm{SG}}$ the noise level, which is recommended by the authors to be in the interval $[0.1, 0.2]$. We set $\alpha_{\mathrm{SG}} = 0.2$ since with this noise level, SmoothGrad produced the explanations with the highest attribution quality in most of the cases in our experiments.

\paragraph{NoiseGrad}
We choose $\sigma_{\mathrm{NG}}$ such that 
the relative accuracy drop $\textrm{AD} (\sigma_{\mathrm{NG}})$ is approximately 5 percent in all experiments. We defined the relative accuracy drop as follows:
\begin{align}
\textrm{AD} (\sigma) = 1 - \frac{\textrm{ACC}(\sigma) - \textrm{ACC}(\infty)}{\textrm{ACC}(0) - \textrm{ACC} (\infty)},
\label{eq:AccuracyDrop}
\end{align}where $\textrm{ACC}(0)$ and $\textrm{ACC}(\infty)$ correspond to the original accuracy and the chance level, respectively.

\paragraph{FusionGrad}

We fix the noise level $\sigma_{\mathrm{SG}}$ for SG by 
Eq. \eqref{eq:NoiseLevelSG} (setting the noise level to $\alpha_{\mathrm{SG}} = 0.1$ i.e., the lower bound of the noise interval as suggested in \cite{smilkov2017smoothgrad}) and adjust the noise level $\sigma_{\mathrm{NG}}$ for NG so that the relative accuracy drop
\begin{align}
\textrm{AD}_{++} (\sigma_{\mathrm{SG}}, \sigma_{\mathrm{NG}}) = 1 - \frac{\textrm{ACC}_{++}(\sigma_{\mathrm{SG}}, \sigma_\mathrm{NG}) - \textrm{ACC}(\infty)}{\textrm{ACC}_{++}(0, 0) - \textrm{ACC} (\infty)}
\label{eq:AccuracyDropPlus}
\end{align}
is again around $\alpha_{\textrm{FG}} = 0.05$. Here,  $\textrm{ACC}_{++} (\sigma_{\mathrm{SG}}, \sigma_{\mathrm{NG}})$ is the classification accuracy with noise levels $\sigma_\mathrm{SG}$ and $ \sigma_\mathrm{NG}$ in the input image and the weight parameters, respectively. 

\begin{figure}[!ht]
\centering
\includegraphics[scale=0.35]{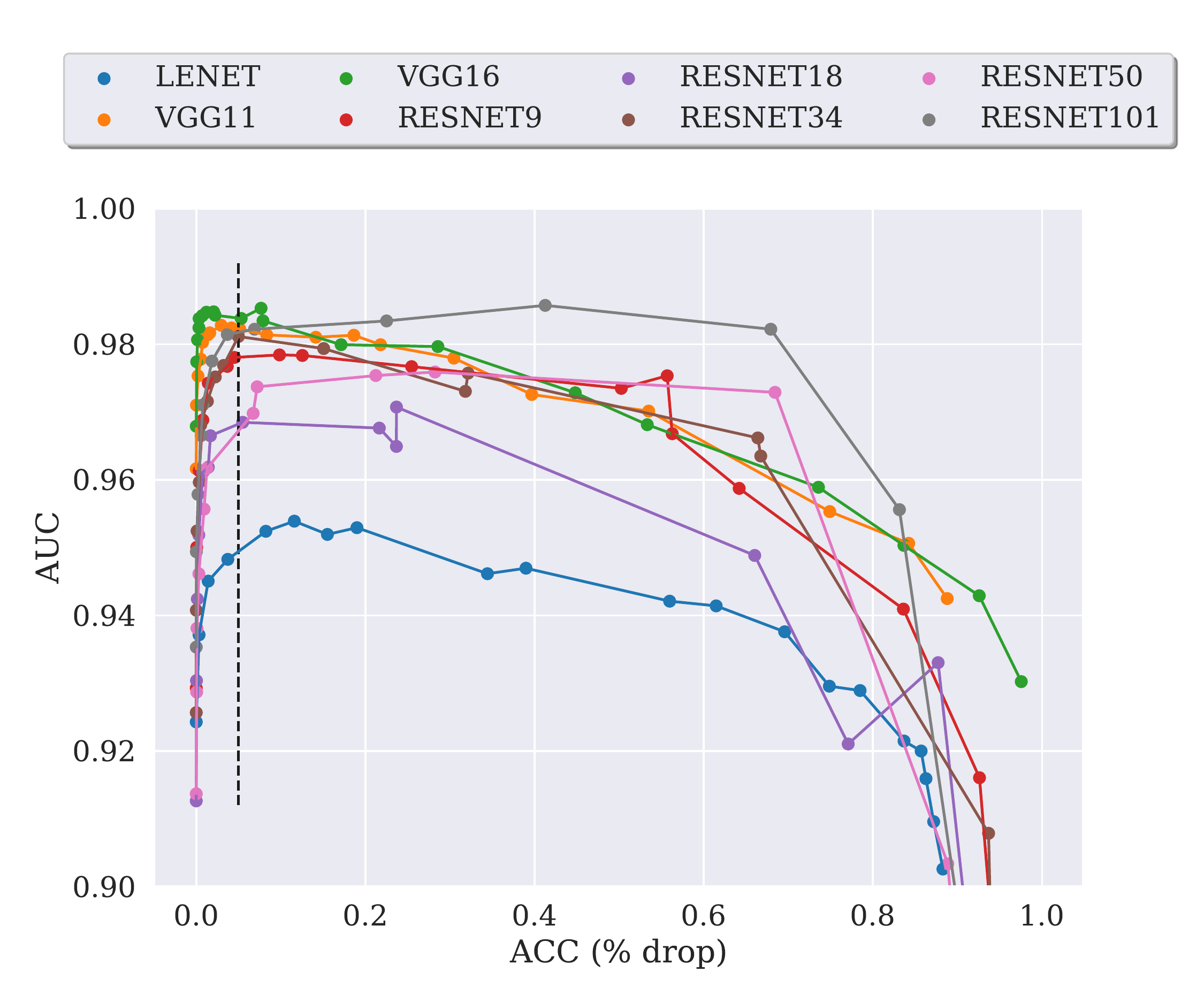} 
\caption{A visual interpretation of the proposed "5 percent accuracy drop" noise heuristic. Each line represents a different model architecture used in the CMNIST experiment and every dot reflects the average results from 200 randomly test samples. We can observe in that general, an increase in the noise level until a drop in accuracy of $\alpha_{\mathrm{NG}} = 0.05$ (black horizontal line), boosts attribution quality (AUC value). }
\label{fig:ACCDROP}
\end{figure}

For a better comprehension of the recommended heuristic, we report the relation between AUC value, i.e., the localization quality of the explanations, and the accuracy drop of the model (ACC) (see Eq. \ref{eq:AccuracyDrop}) in Figure \ref{fig:ACCDROP}. The more noise added to the weights, the higher is the accuracy drop, which is reported on the x-axis. Interestingly, we observe an increase in the AUC value, reported on the y-axis, until an accuracy drop of 5 percent. Adding more noise to the weights leads to a greater accuracy drop and causes the AUC values to decrease again.

\subsection{Connection to Diagonal and KFAC Laplace approximation}

As described in the main manuscript, NoiseGrad with multiplicative Gaussian noise can be seen as performing a quite crude Laplace Approximation. Yet, despite the simplicity of the proposed method, it is still sufficiently accurate to get an insight into the uncertainty of the model to enhance explanations. 

\begin{figure*}[!ht] 
\centering
\includegraphics[scale=0.15]{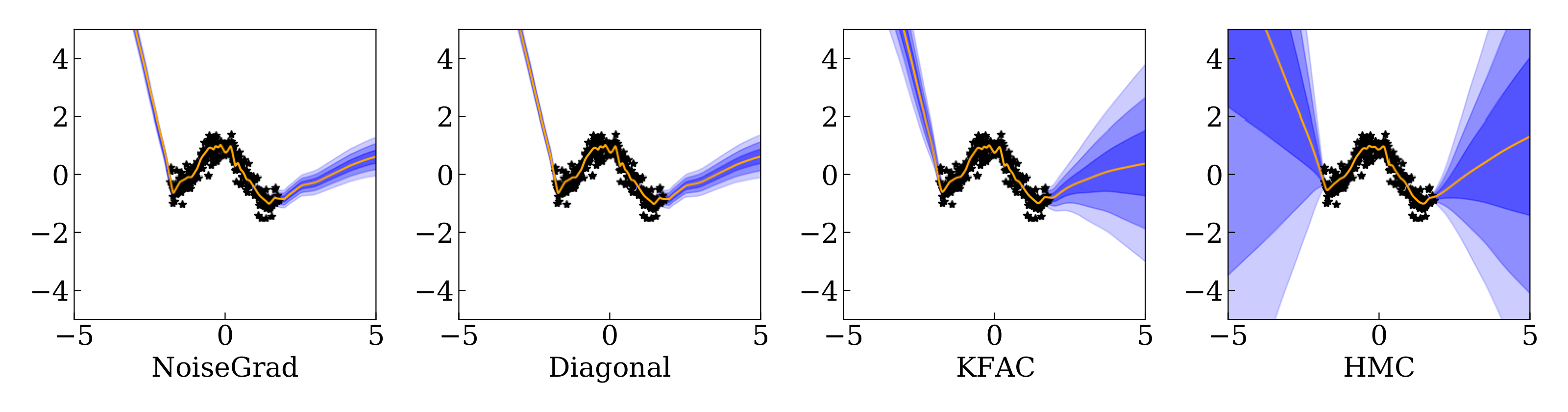}
\caption{Toy example: comparison of NoiseGrad with Diagonal, KFAC and Hamiltonian Monte Carlo Approximation. Parameters for the NoiseGrad, Diagonal and KFAC Laplace approximations were chosen, such that they have a similar performance (MSE loss) on thee test dataset.}
\label{fig:toyExample}
\end{figure*}

We illustrate uncertainties obtained by our proposed method on a toy-task -- a one-dimensional regression generated by a Gaussian Process with RBF kernel with parameters $l = 1$, $\sigma= 0.3$. In total we generated $400$ points, $250$ were used for training, and $150$ for testing. For training, we used a simple two-layer MLP with $10$ hidden neurons respectively and sampled $1000$ models using Hamiltonian Monte Carlo \cite{neal1993bayesian} using the \texttt{hamiltorch} library \cite{cobb2020scaling}.  The sampled model used to perform both Kroneker-Factored (KF) and Diagonal Laplace Approximations using the $\texttt{curvature}$ library \cite{lee2020estimating}. Finally, we visualize the uncertainty of the predictions by adding multiplicative noise to the weights using NoiseGrad, comparing it to the KFAC and Diagonal Laplace approximations. Hyperparameters for all three methods were chosen to have approximately the same loss on the test set. All results are shown in Figure \ref{fig:toyExample}, which demonstrates that NoiseGrad has not quite as precise, but a similar estimate of the model uncertainty as to the Bayesian approximation methods.

\begin{figure*}
\centering
\includegraphics[scale=0.15]{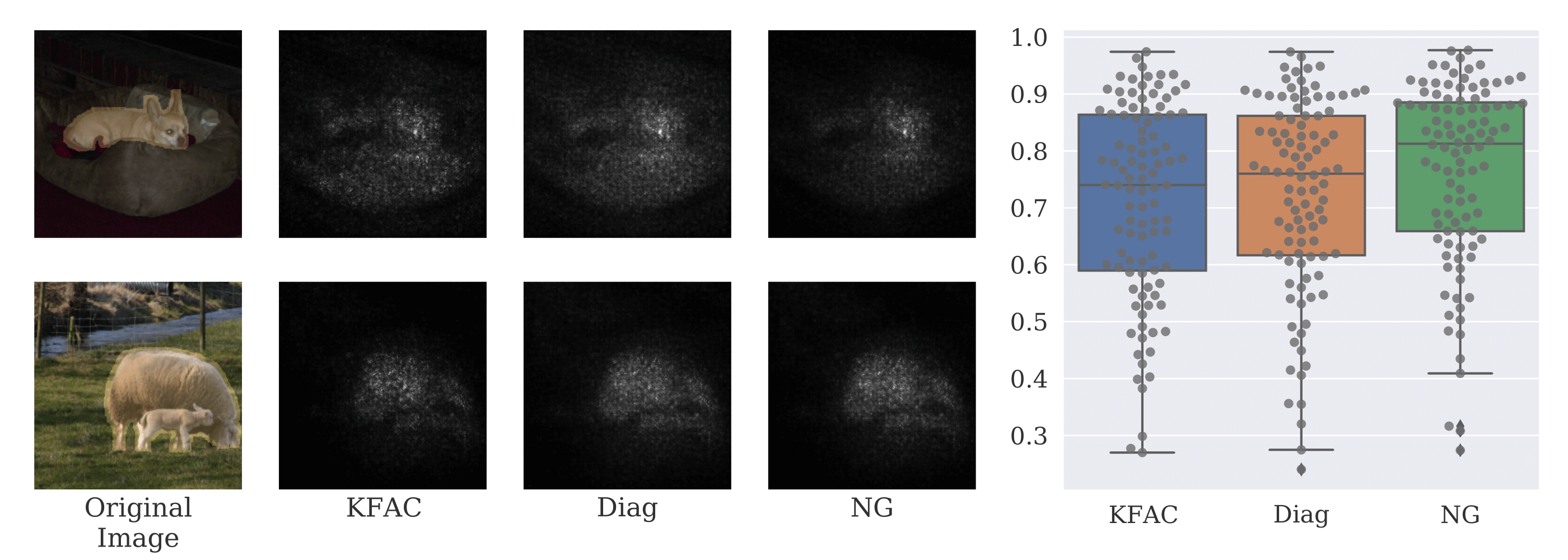}
\caption{Comparison of the performance between our proposed method NG, KFAC and Diagonal Laplace Approximation. We can observe that the attribution quality (left: for 2 randomly chosen images) is comparable both visually and quantitatively (right: measured by $\mathrm{AUC}$) for the different approximation methods.}
\label{fig:BayesianLearingConnection}
\end{figure*}

Empirically, from Figure \ref{fig:BayesianLearingConnection}, we can observe that explanations, which were enhanced by NoiseGrad have only slight visual differences compared to the explanations obtained by proper Laplace Approximation methods, i.e., Diagonal or KFAC. For this experiment we performed Laplace Approximation on ResNet-18, trained for multi-label classification on the PASCAL VOC 2012 dataset. Setting the NoiseGrad hyperparameter $\sigma$ to $0.1$ we searched for the hyperparameters of Diagonal and KFAC Laplace approximation such that all models had on average a similar performance on the test data.

\subsection{Experiments}

In the following, we will provide additional information regarding the experimental setup as described in the main manuscript. All experiments were computed on Tesla P100-PCIE-16GB GPUs. 

\subsubsection{Evaluation methodology} 

In the main manuscript, we presented four properties considered useful for an explanation to fulfill. To measure the extent to which the explanations satisfied the properties, empirical interpretations are necessary. In the following paragraphs, we thus motivate and discuss choices made with respect to evaluation.

Having the object segmentation masks available for the input images, we can formulate the first property as follows

\vspace{0.25cm}
\textit{\textbf{(Property 1) Localization} measures the extent to which an explanation attributes its explainable evidence to the object of interest.}
\vspace{0.25cm}

The higher the concentration of attribution mass on the ground-truth mask the better. For this purpose, there exist many metrics in the literature \cite{ theiner2021interpretable, bach2015analyzing, kohlbrenner2020best, zhang2016topdown} that can be applied. Since we are interested in the attributions' rankings and in particular, in the set of attributions covered by the ground truth (GT) mask, we apply the Relevance Rank Accuracy \cite{arras2021ground} as a single metric in order to assess the explanation localization. This metric computes the ratio of high-intensity relevances (or attributions) that lie within the ground truth mask. Scores range [0, 1] where higher scores are preferred. Localization is defined as follows

$$
RA(E_{topK}, GT) = \frac{\mid{E_{topK}} \cap GT \mid}{\mid GT \mid}
$$

where $topK$ is the size of the ground truth mask and $E_{top K} = \{i_1, i_2, ..., i_l \mid R_{i1} \leq R_{i2} ... \leq R_{iK}\}$ \cite{arras2021ground}. The metric counts how many of the top features lie within the ground truth mask to then divide by the size of the mask.

The second property attempts to understand whether the assigned attributions accurately reflect the behavior of the model, i.e., the faithfulness of the explanations. It is arguably one of the most well-studied evaluation metrics, hence many empirical interpretations have been proposed \cite{bach2015pixel, montavon2018methods, samek2016evaluating, yeh2019fidelity, nguyen2020quantitative, alvarez2018towards, rieger2020simple}. We define it as follows

\vspace{0.25cm}
\textit{\textbf{(Property 2) Faithfulness} estimates how the presence (or absence) of features influence the prediction score: removing highly important features results in model performance degradation. }
\vspace{0.25cm}

To evaluate the relative fulfillment of this property, we conduct an experiment that iteratively modifies an image to measure the correlation between the sum of attributions for each modified subset of features and the difference in the prediction score \cite{bhatt2020evaluating}. Given a model $f$, an explanation function $E$ and a subset $|S|$ of $d$ indices of samples $\boldsymbol{x}$ and a baseline value $\overline{{x}}$, we define {faithfulness} as follows 

$$
 \mu_{\mathrm{F}}({E}, {f}; \boldsymbol{x}) = \underset{S \in\big(\underset{|S|}{[d]}\big)}{\operatorname{corr}}\left(\sum_{i \in S} {E}({f}, {x})_{i}, {f}({x})-{f}\left(\boldsymbol{x}_{\left[{x}_{s}=\overline{{x}}_{s}\right]}\right)\right)
$$

Due to the potential appearance of spurious correlations and creation of out-of-distribution samples, while masking original input \cite{hooker2019benchmark}, the choice of the pixel perturbation strategy is non-trivial \cite{sundararajan2017axiomatic, sturmfels2020visualizing}. For each test sample, we, therefore, average results over 100 iterations, using a subset size $|S| = 32$ while setting the baseline value $\overline{{x}}$ to black. In an attempt to capture linear dependencies, the correlation metric is set to Pearson \cite{bhatt2020evaluating}.

In the third property to assess attribution quality, we examine the robustness of the explanation function \cite{kindermans2019reliability, alvarez2018towards} which, with subtle variations also is referred to as \textit{continuity} \cite{montavon2018methods}, \textit{stability} \cite{alvarez2018towards}, \textit{coherence} \cite{guidotti2020black}.

\vspace{0.25cm}
\textit{\textbf{(Property 3) Robustness} measures how strongly the explanations vary within a small local neighborhood of the input while the model prediction remains approximately the same.}
\vspace{0.25cm}

 Our empirical interpretation of robustness follows \cite{yeh2019fidelity} definition of \textit{max-sensitivity} which is a metric that measures maximum sensitivity of an explanation when the subject is under slight perturbations, using Monte Carlo sampling-based approximation. It is defined as follows
 
$$
\operatorname{SENS_{\operatorname{MAX}}}(E, f, {x}, r) =\max _{\|{x'}-{x}\| \leqslant r} \| E({f}, {x'})-E({f}, x) \|
$$

It works by constructing $n$ perturbed versions of the input $\boldsymbol{x'} = \boldsymbol{x} + \delta$ by sampling uniformly random from a $L_{\infty}$ ball over the input, where radius=0.2 and $n$=10. By subtracting the Frobenius norms of $E(\boldsymbol{x})$ from $E(\boldsymbol{\boldsymbol{x'}})$ we can measure the extent to which the explanations of $\boldsymbol{x}$ and $\boldsymbol{x'}$ remained close (or distant) under these perturbations. The maximum value is computed over this set of $n$ perturbed samples. A lower value for \textit{max-sensitivity} is considered better.

In the final property to assess attribution quality we were interested in studying the sparseness of the explanations since if the number of influencing features in an explanation is too large, it may be too difficult for the user to understand the explanation \cite{nguyen2020quantitative}. In efforts to make \textit{human-friendly} explanations, sparseness might therefore be desired.  In this category of metrics, there are also \textit{complexity} \cite{bhatt2020evaluating} and \textit{effective complexity} \cite{nguyen2020quantitative} metrics. 

\vspace{0.25cm}
\textit{\textbf{(Property 4) Sparseness} measures the extent to which features that are only truly predictive of the model output have significant attributions.}
\vspace{0.25cm}

Following \cite{chalasani2020concise}, we define sparseness as the Gini index \cite{hurley2009comparing} of the absolute value of the attribution vector $E(\boldsymbol{x})$ where $n$ is the length of the attribution vector.

$$
G(E) = \frac{\sum_{i=1}^{n}(2 i-n-1) E_{i}}{n \sum_{i=1}^{n} E_{i}}
$$

\begin{figure*}
\centering
\includegraphics[scale=0.11]{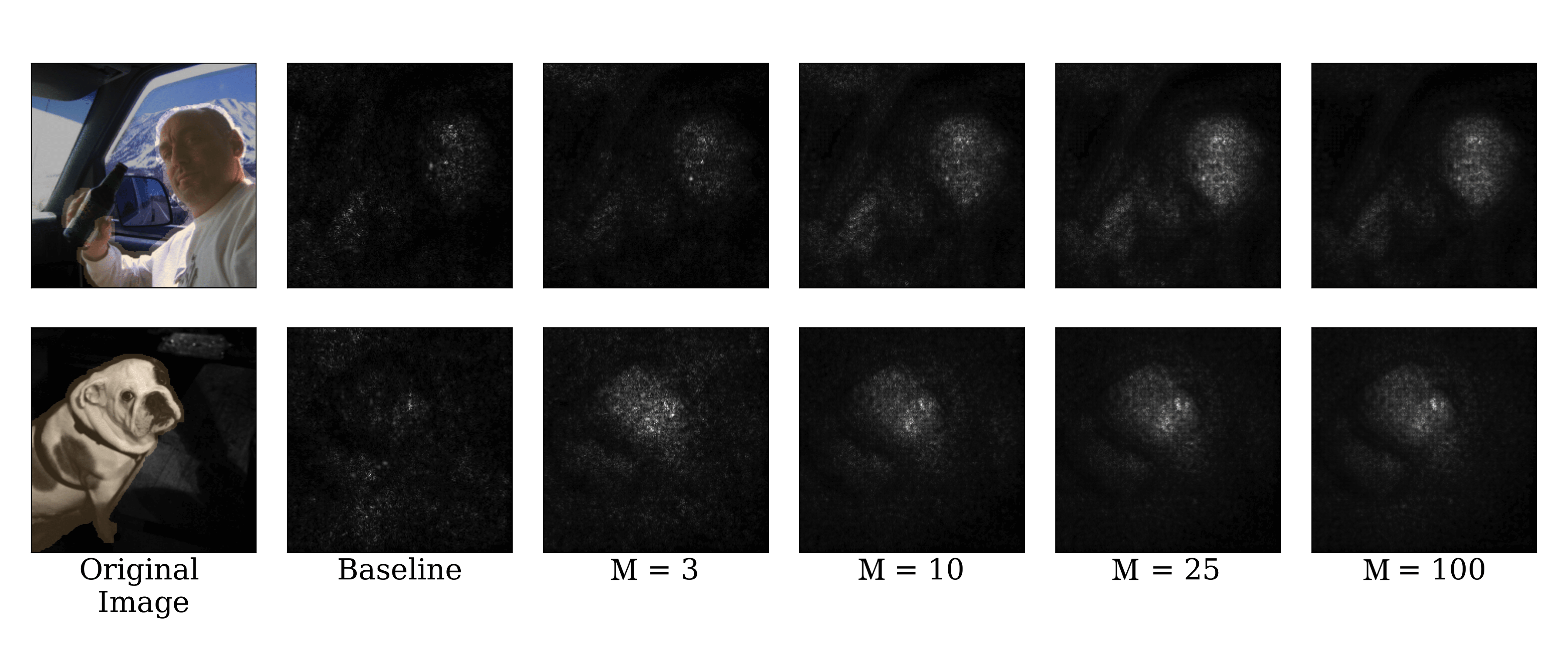}
\caption{Saliency explanations for two randomly chosen PASCAL VOC 2012 images using NoiseGrad with different sample sizes. The quality of the explanation improves with an increasing number of samples until it no longer changes when the number of samples reaches 25.}
\label{fig:samplesize}
\end{figure*}

\subsection{Datasets and pre-processing}

In our experiments, we use two datasets where pixel-wise placements of the explainable evidence of the attributions were made available.

\paragraph{CMNIST} For constructing the semi-natural dataset, we combined two well known datasets; the MNIST dataset \cite{lecun2010mnist}, consisting of $60,000$ train- and $10,000$ test grey-scale images and the CIFAR-10 dataset, consisting of $50,000$ train- and $10,000$ test colored images. Each MNIST digit (0 to 9) was first resized from $(28, 28)$ to $(16, 16)$ and then rotated randomly, with a maximal rotation degree of 15. Afterward, those digits were added on top of CIFAR images of size $(32, 32)$, which were uniformly sampled, such that there exists no correlation between CIFAR and MNIST classes. Hence, within this experimental setup, we could assume that the model does not rely on the background for the decision-making process. Using this constructed semi-natural dataset in the experiments, random noise of varying variance $\sigma^2$ was added to each CMNIST image. 

As a pre-processing step, all images were normalized. For the training set, the random affine transformation was also applied to keep the image center invariant. The random rotation ranges withing $[-15, 15]$ degrees, random translations, scaling and shearing in the range of $[0.95, 1]$ and $[0, 0.05]$ percents respectively using the fill color $0$ for standard \texttt{torchvision.transforms.RandomAffine} data augmentation in \texttt{PyTorch} \cite{NEURIPS2019_9015}. We constructed three different ground-truth segmentation masks for the CMNIST dataset as follows: squared box around the digit, the digit segmentation itself, and the digit plus neighborhood of 3 pixels in each direction. In the experiments, as the evidence for a class typically also is distributed around the object itself, the squared box of $(16, 16)$ pixels was selected as the single ground-truth segmentation.

\paragraph{PASCAL VOC 2012} We use PASCAL Visual Object Classes Challenge 2012 (VOC2012) \cite{Everingham10} to evaluate our method on a natural high-dimensional segmented dataset. For training and evaluation, we resize images such that the number of pixels on one axis is 224 pixels and perform center crop afterward, resulting in square images of $224 \times 224$ pixels. Images are then normalized with standard ImageNet mean and standard deviation. The segmentations are also resized and cropped, accordingly. For the training we augment the data with random affine transformations that include random rotation in the range $[-30, 30]$ degrees, random translations, scaling, and shearing in the range of $[0, 5]$. Additionally, we perform horizontal and vertical random flips with a probability of $0.5.$

\paragraph{ILSVRC-15} We additionally used ImageNet Large Scale Visual Recognition Challenge (ILSVRC-15) \cite{ILSVRC15} to evaluate the different explanation-enhancing methods. We applied the same pre-processing as outlined for the VOC2012 dataset apart from the random affine transformations and scaling that was used for training purposes - instead, we used a ResNet-18 classifier with pre-trained weights.

\subsubsection{Explanation methods}
\label{appendix:appA}

In this section, we provide a brief overview of the explanations methods used in the experiments conducted in both the main manuscript and the supplementary material. For each explanation method, we use absolute values of the attribution maps, following the same logic as in \cite{smilkov2017smoothgrad}. This is motivated by the fact that both positive and negative attributions of explanations drive the final prediction score.

\paragraph{Saliency} Saliency (SA) \cite{morch} is used as the baseline explanation method. The final relevance map quantifies the possible effects of small changes in the input on the output produced $$E_{Saliency}\left(\boldsymbol{x}, f(\cdot, W)\right) = \frac{\partial f(\boldsymbol{x}, W)}{\partial \boldsymbol{x}}.$$
    
\paragraph{Integrated Gradients}
     Integrated Gradients (IG) \cite{sundararajan2017axiomatic} is an axiomatic local explanation algorithm that also addresses the "gradient saturation" problem. A relevance score is assigned to each feature by approximating the integral of the gradients of the models' output with respect to a scaled version of the input \cite{sundararajan2017axiomatic}. The relevance attribution function for IG is defined as follows
    
    $$E_{IG}\left(\boldsymbol{x}, f(\cdot, W)\right) =  \left(\boldsymbol{x} - \bar{\boldsymbol{x}} \right) \int_0^1 \frac{\partial f (\boldsymbol{x} + \alpha(\boldsymbol{x} - \bar{\boldsymbol{x}}), W)}{\partial \boldsymbol{x}} d\alpha, $$
    where $\bar{x}$ is a \textit{reference point}, which is chosen in a way that it represents the absence of a feature in the input.

\paragraph{GradShap}
    Gradient Shap (GS) \cite{lundberg2017unified} is a local explainability method that approximates Shapley values of the input features by computing the expected values of the gradients when adding Gaussian noise to each input. Since it computes the expectations of gradients using different reference points, it can be viewed as an approximation of IG.

\paragraph{Occlusion.}
  The occlusion method proposed in \cite{zeiler2014visualizing} is a local, model-agnostic explanation method that attributed relevances to the input by masking specific regions of the input and collecting the information about the corresponding change of the output.

\paragraph{LRP} Layer-wise Relevance Propagation \cite{bach2015pixel} is a model-aware explanation technique that can be applied to feed-forward neural networks and can be used for different types of inputs, such as images, videos, or text \cite{anders2019understanding, arras2017relevant}. Intuitively, the core idea of the LRP algorithm lies in the redistribution of a prediction score of a certain class towards the input features, proportionally to the contribution of each input feature. More precisely, this is done by using the weights in combination with the neural activations that were generated within the forward pass as a measure of the relevance distribution from the previous layer to the next layer until the input layer is reached. Note that this propagation procedure is subject to a conservation rule -- analogous to Kirchoff’s conservation laws in electrical circuits \cite{montavon2019layer} -- meaning, that in each step of back-propagation of the relevances from the output layer towards the input layer, the sum of relevances stays the same. 

In the experiments we deploy the LRP gamma rule i.e., LRP-$\gamma$ which is dependent on the tuning of $\gamma$ can favor the positive attributions over negative ones. As we increase $\gamma$, negative contributions start to cancel out. The LRP gamma rule is defined as follows  
$$
R_{j}=\sum_{k} \frac{a_{j} \cdot\left(w_{j k}+\gamma w_{j k}^{+}\right)}{\sum_{0, j} a_{j} \cdot\left(w_{j k}+\gamma w_{j k}^{+}\right)} R_{k}
$$
where $\gamma$ controls how much positive contributions are favoured, $j$ and $k$ are neurons at two consecutive layers of the network, $a_j$ are attributions, $w$ are weights aand $R_{k}$ are relevances.

\subsubsection{Additional results} 

In the following, we present results from additional experiments.

\begin{figure*}
\centering
\includegraphics[width=.7\linewidth]{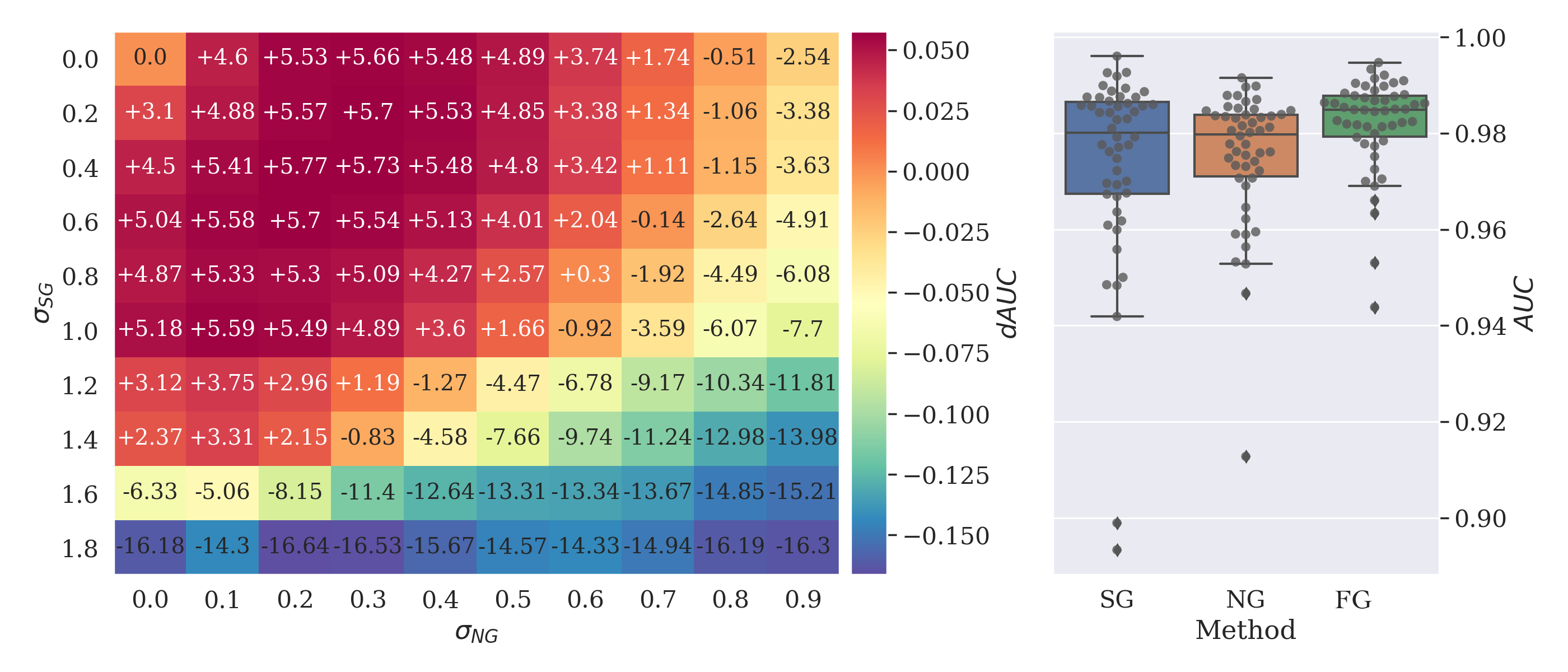}
\caption{Left: Noise level dependence of relative AUC improvement dAUC from the Baseline method. The horizontal and the vertical axis corresponds to the noise levels $\sigma_{NG}$ and $\sigma_{SG}$. Right: the relative AUC improvement with optimal hyperparameter choice. We can observe that FG achieves better performance than SG or NG alone.}
  \label{figgridsearch}
\end{figure*}

\paragraph{Dependence on noise level} 
In the following, we investigate how the ranking of attributions in terms of the area under the curve ($\mathrm{AUC}$) \cite{10.1016/j.patrec.2005.10.010} depends on the noise levels, $\sigma_{\mathrm{SG}}$ and $\sigma_{\mathrm{NG}}$, for FusionGrad.  Note that FusionGrad with $\sigma_{\mathrm{NG}} = 0$ and
$\sigma_{\mathrm{SG}} = 0$, respectively, correspond to SmoothGrad and NoiseGrad.
Figure \ref{figgridsearch} (left) shows 
\emph{relative AUC improvement} $d\mathrm{AUC} = (\mathrm{AUC}$ / $\mathrm{AUC}_{\mathrm{Baseline}}) -1)$ to the Baseline method (with zero noise levels).

From Figure \ref{figgridsearch} we can observe that the heuristics for SG, NG, and FG represents the noise levels that give the highest attribution quality in terms of ranking, which we empirically interpreted as $\mathrm{AUC}$.
We also investigate the potential of our proposed methods when the noise levels are tuned by optimizing the ranking criterion.
Figure \ref{figgridsearch} (right) plots the $\mathrm{AUC}$s with the optimal noise levels over 200 test samples.
We observe that stochasticity in the input space and the parameter space potentially achieve the same level of attribution ranking (with smaller variance with the latter) and that their combination can further boost the performance.

\paragraph{Extended table 3 - additional explanation methods}

To understand how the different explanation-enhancing methods SG, NG, and FG compare to our given baseline, we scored the explanations using 4 criteria of attribution quality. The scores were computed and averaged over 256 randomly chosen test samples, using a ResNet-9 classifier. In the main manuscript, we reported on Saliency \cite{morch} explanations. In the extended Table 3 below, results for Integrated Gradients \cite{sundararajan2017axiomatic} and Gradient Shap \cite{lundberg2017unified} are included. 

As seen in Table \ref{sample-table2}, FusionGrad is significantly better than both SmoothGrad and Baseline in all evaluation criteria --- that is, FusionGrad is producing more localized, faithful, robust, and sparse explanations than those baselines and competing approaches for the Integrated Gradient explanation. NoiseGrad is producing slightly higher scores than FusionGrad in robustness and faithfulness criteria, but as shown in the table, these results are not significant. 
Further, for Gradient Shap explanations, we can observe that in general, NoiseGrad and FusionGrad outperform the baseline- and comparison methods in this setting as well.

\begin{table*}[h!]
  \caption{Comparison of attribution quality on different explanation methods for CMNIST dataset, where the noise levels are set by the heuristic.  $\uparrow$ and $\downarrow$ indicate the larger is the better and the smaller is the better, respectively. 
The values of the best method and the methods that are not significantly outperformed by the best method,
according to the Wilcoxon signed-rank test for $p=0.05$,
are bold-faced. 
  }
  \label{sample-table2}
  \centering
  \begin{tabular}{llcccc}
    \toprule
    {Explanation}     &  {Method}     & {Localization ($\uparrow$)}    & {Faithfulness ($\uparrow$)}  &  {Robustness ($\downarrow$)} & {Sparseness ($\uparrow$)}  \\
    \midrule

Integrated Gradients & Baseline 	 & 0.6400 $\pm$  0.0582  & 0.3430 $\pm$  0.1481  & 0.0556 $\pm$  0.0163  & 0.7641 $\pm$  0.0476 \\
& SG 	 & 0.6679 $\pm$  0.0572 & 0.3506 $\pm$  0.1558 & 0.0585 $\pm$  0.0211  & 0.7750 $\pm$  0.0438 \\
& NG 	 & 0.6883 $\pm$  0.0522 & \textbf{0.3593} $\pm$  0.1555  & \textbf{0.0170} $\pm$  0.0048  & 0.8102 $\pm$  0.0371 \\
& FG 	 & \textbf{0.6981} $\pm$  \textbf{0.0511}  	 & \textbf{0.3537} $\pm$  \textbf{0.1556} &  \textbf{0.0179} $\pm$  \textbf{0.0049} & \textbf{0.8199} $\pm$  \textbf{0.0354}  \\

\midrule
Gradient Shap & Baseline 	 & 0.4522 $\pm$  0.0438 & 0.3623 $\pm$  0.1625 & 0.1965 $\pm$  0.0640  & 0.7780 $\pm$  0.0479 \\
& SG 	 & 0.4234 $\pm$  0.0398 & \textbf{0.3629} $\pm$  \textbf{0.1535} & 0.1527 $\pm$  0.0538  & 0.7792 $\pm$  0.0424 \\
& NG 	 & \textbf{0.4707} $\pm$  \textbf{0.0430} & \textbf{0.3650} $\pm$  \textbf{0.1589} & \textbf{0.1456} $\pm$  \textbf{0.0486} & 0.8095 $\pm$  0.0378 \\
& FG 	 & 0.4626 $\pm$  0.0437 & 0.3603 $\pm$  0.1536 & \textbf{0.1464} $\pm$  \textbf{0.0481} & \textbf{0.8200} $\pm$  \textbf{0.0352}  \\

    \bottomrule
  \end{tabular}
\end{table*}

\paragraph{Extended table 4 - additional datasets}

To understand how the benefits of proposed explanation-enhancing methods NoiseGrad and FusionGrad extend to other datasets, we evaluated all the approaches using the ILSVRC-15 dataset \cite{ILSVRC15}. The scores were computed for all four attribution quality criteria and averaged over 250 randomly chosen, normalized test samples, using a ResNet-18 classifier with pre-trained weights. As seen in the extended Table \ref{imagenet},  explanations enhanced by FusionGrad or NoiseGrad are in most criteria outperforming both SmoothGrad and Baseline --- especially when it comes to making explanations more faithful and localized. Results from three explanation methods i.e., Saliency \cite{morch}, Integrated Gradients \cite{sundararajan2017axiomatic} and Gradient Shap \cite{lundberg2017unified} are included in the table.

\begin{table*}[h!]
  \caption{Comparison of attribution quality on different explanation methods for ImageNet dataset, using noise levels $\sigma_{\mathrm{NG}}=0.2$, $\sigma_{\mathrm{SG}}=0.4758$ and $\sigma_{\mathrm{SF}}=(\sigma_{\mathrm{NG}}/2, \sigma_{\mathrm{SG}}/2)$.  $\uparrow$ and $\downarrow$ indicates the larger is the better and the smaller is the better, respectively. The values of the best method and the methods that are not significantly outperformed by the best method,
according to the Wilcoxon signed-rank test for $p=0.05$,
are bold-faced. 
  }
  \label{imagenet}
  \centering
  \begin{tabular}{llcccc}
    \toprule
    {Explanation}     &  {Method}     & {Localization ($\uparrow$)}    & {Faithfulness ($\uparrow$)}  &  {Robustness ($\downarrow$)} & {Sparseness ($\uparrow$)}  \\
    \midrule

Saliency 
&
Baseline 	 &
0.5479 $\pm$  0.2526  &
0.0209 $\pm$  0.0660  &
0.0275 $\pm$  0.0134  &
\textbf{0.4329}  $\pm$  \textbf{0.0379} \\
& 
SG 	 
&
0.6881 $\pm$  0.1875 &
0.0330 $\pm$  0.0683  &
0.0178 $\pm$  0.0080  &
0.2306 $\pm$  0.0529 \\
&
NG 	 
&
0.6845 $\pm$  0.1849 &
0.0340 $\pm$  0.0682 & 
0.0160 $\pm$  0.0073  &
0.2951 $\pm$ 0.0543 \\
& FG 	 
&
\textbf{0.7466} $\pm$  \textbf{0.1438}  	 &
\textbf{0.0405} $\pm$  \textbf{0.0710} &
\textbf{0.0036} $\pm$  \textbf{0.0018} &
0.2128 $\pm$  0.0488  \\

\midrule
Integrated Gradients &
Baseline 	 
&
0.5273 $\pm$  0.2448  &
0.3430 $\pm$  0.1481  &
0.0373 $\pm$  0.0134  &
0.5718 $\pm$  0.0379 \\
& SG 	 
&
0.5428 $\pm$  0.1919 &
0.3506 $\pm$  0.1558 &
0.0084 $\pm$  0.0023  &
0.5378 $\pm$  0.0333 \\
&
NG 	 &
\textbf{0.5968} $\pm$  \textbf{0.1919} &
\textbf{0.3593} $\pm$  \textbf{0.1555}  &
0.0095 $\pm$  0.0026  &
0.\textbf{6013} $\pm$  \textbf{0.0486} \\
&
FG 	 &
0.5728 $\pm$  0.2072  	 &
\textbf{0.3537} $\pm$  \textbf{0.1556} &
\textbf{0.0036} $\pm$  \textbf{0.0007} &
0.5548 $\pm$  0.0415 \\

\midrule
Gradient Shap 
&
Baseline 	 &
\textbf{0.5773} $\pm$  \textbf{0.2185} &
0.0145 $\pm$  0.0613 &
0.0384 $\pm$  0.0152  &
\textbf{0.6111} $\pm$  \textbf{0.0547} \\
& SG 	 
&
0.4778 $\pm$  0.2693 &
0.0136 $\pm$  0.0599 &
\textbf{0.0019} $\pm$  \textbf{0.0006}  &
0.5170 $\pm$  0.0273 \\
& NG 	 
& 
\textbf{0.5913} $\pm$  \textbf{0.1957} &
0.0183 $\pm$  0.0625 &
0.0077 $\pm$  0.0021 &
0.5957 $\pm$  0.0421 \\
& FG 	 
&
0.5694 $\pm$  0.2107 &
\textbf{0.0195} $\pm$  \textbf{0.0642} &
0.0040 $\pm$  0.0007 &
0.5502 $\pm$  0.0421  \\

    \bottomrule
  \end{tabular}
\end{table*}

\paragraph{Sanity checks - model parameter randomization tests}

In addition to the existing evaluation criteria, we also tested the relative sensitivity of the explanation-enhancing methods to the parameterization of the models using the methodology outlined in \cite{adebayo2018sanity}. Experimental results positively suggest that in general --- explanations for the different approaches are not completely uninformative on the trained weights i.e., the rank correlation coefficient between explanations of a randomized- versus a non-randomized model $\neq 1$. The correlation coefficient is slightly higher for NG/ FG compared to SG which is due to the multiplicative noise that is added to the model weights.

\subsubsection{Models and training results}

Each model was trained similarly; using a standard cross-entropy loss (\texttt{torch.nn.CrossEntropyLoss} in \texttt{PyTorch}) -- the LeNet and ResNet networks were trained using the AdamW variant \cite{loshchilov2019decoupled}, with an initial learning rate of $0.001$ and the VGGs were trained using stochastic gradient descent with an initial learning rate of $0.01$ and momentum of $0.9$. The training was completed for $20$ epochs for all models. Furthermore, ReLU activations were used for all layers of the networks apart from the last layer, which employed a linear transformation. For details on model architectures we point the reader to the original source; LeNet \cite{lecun1998gradient}, ResNets \cite{he2015deep} and VGG \cite{simonyan2014very} respectively. 

In Table \ref{table-cmnist-accs}, the reader can find the model performance listed (both training- and test accuracies). For the sake of space, we refer to each ResNet as RN.

\begin{table*}[!ht]
  \caption{Training results for different architectures}
  \label{table-cmnist-accs}
  \centering
  \begin{tabular}{lcccccccc}
    \toprule
    Acc. & LeNet & VGG11  & VGG16  & RN9 & RN18 & RN34 & RN50 & RN101  \\ 
    \midrule
    Train & 0.7841 & 0.9237 & 0.9508 & 0.9723 & 0.9747 & 0.9734 & 0.9754 & 0.9760 \\
    Test & 0.8653 & 0.9634 & 0.9797 & 0.9850 & 0.9864 & 0.9868 & 0.9898 & 0.9885 \\
    \bottomrule
  \end{tabular}
\end{table*}

In the VOC2012 dataset, we trained a ResNet-18 for multi-label classification, changing the number of output neurons to 20 to represent the number of classes in the dataset. The network was trained with Binary Cross-Entropy loss applied on top of sigmoid layer (\texttt{torch.nn.BCEWithLogitsLoss} in \texttt{PyTorch}). ResNet-18 was initialized with ImageNet \cite{deng2009imagenet} pre-trained weights (available in \texttt{Torchvision.models}) and was trained for 100 epochs with SGD with $0.01$ learning rate and with 100 epochs with a learning rate of $0.005$. The network achieves a loss of $0.000295$ on the training dataset and $0.000376$ on the test dataset. More illustrations could be found in the Appendix.

\paragraph{Choosing $N$ for NG and FG} Proposed NG and FG methods can be seen as a version of a Monte-Carlo integration \cite{metropolis1949monte} for the following integrals:

$$I_{NG}(\boldsymbol{x}) = \int_{\mathbb{R}^{S}}  E\left(\boldsymbol{x}, f(\cdot,\mathcal{W})\right)p(\mathcal{W})d\mathcal{W},$$

$$I_{FG}(\boldsymbol{x}) = \int_{\mathbb{R}^{S}}\int_{\mathbb{R}^{d}}  E\left(\boldsymbol{x} + \xi, f(\cdot, \mathcal{W}_i)\right)p(\xi)p(\mathcal{W})d\xi d\mathcal{W}. $$
\vspace{0.25cm}

As a Monte-Carlo approximation, the standard error of the mean decreases asymptotically as $\frac{1}{\sqrt{M}}$ is independent of the dimensionality of the integral. In practice, we observe that a sample size $M \in [25,50]$ is already sufficient to generate appealing explanations as shown in Figure \ref{fig:samplesize}. For FG, only 10 samples for both $M$ (NG samples) and $N$ (SG samples) are enough to enhance explanations. 

\subsection{Related Works}

While adding an element of stochasticity during back-propagation to improve generalization and robustness is nothing new in the Machine Learning community \cite{an1996effects, Gal16, poole2014analyzing, blundell2015weight} -- previously, to the best of our knowledge, it was not used in the context of XAI before.

Most similar to our proposed NoiseGrad method are approaches that also inject stochasticity at inference time e.g., \cite{Gal16} which we also do. Our method can further be associated with \textit{ensemble modeling} \cite{dietterich2000ensemble}, where an ensemble of predictions from several classifiers are averaged, to increase the representation of the model space or simply, reduce the risk of choosing a sub-optimal classifier. Ensemble modeling is motivated by the idea of "\textit{wisdom of the crowd}", where a decision is based on the collective opinion of several experts (models) rather than relying on a single expert opinion (single model). In our proposed method, by averaging over a sufficiently large number of samples $N$, the idea is that the noise associated with each explanation will be approximately eliminated. 

SmoothGrad \cite{smilkov2017smoothgrad} -- being the main comparative method of this work -- has gained a great amount of popularity since its publication. That being said, to our awareness, very few additions have been put forward when it comes to extending the idea of averaging explanations from noisy input samples. However, in the simple operation of \textit{averaging} to enhance explanations, a few ideas have been proposed. The work by \cite{bhatt2020evaluating} shows that the typical error of aggregating explanation functions is less than the expected error of an explanation function alone and \cite{rieger2020simple} suggests a simple technique of averaging multiple explanation methods to improve robustness against manipulation.

\subsection{Broader Impact}

Although the use of machine learning algorithms in various application areas, such as autonomous driving or cancer cell detection, has produced astonishing achievements, the risk remains that these algorithms will make incorrect predictions. This poses a threat, especially in safety-critical areas as well as raises legal, social, and ethical questions. 
However, in times of advancing digitization, the deployment of artificial intelligence (AI) is often decisive for competitiveness. Therefore, methods for a thorough understanding of the often highly complex AI models are indispensable. This is where the field of explainable AI (XAI) has established itself and in recent years various XAI methods for explaining the "black-box models" have been proposed. 
In general, more precise and reliable explanations of AI models would have a crucial contribution to ethical, legal, and economic requirements, the decision-making basis of the developer, and the acceptance by the end-user.

In particular, it has been shown with the explanation-enhancing SmoothGrad method, that small uncertainties, which are artificially added to the input space can lead to an improvement in the explanation. In this work, we propose a novel method, where the uncertainties are not introduced to the input space, but to the parameter space of the model. We were able to show in various experiments that this leads to a significant improvement when it comes to attribution quality of explanations, measured with different metrics, and subsequently can be beneficial in practice. 
In case the proposed method does not provide a localized, robust, and faithful explanation, this would have no direct consequences for the user as long as the explanation is used for decision support and is not a separate decision unit. The data we use do not contain any personal data or any offensive content.

\subsection{NoiseGrad and Local Explanations}
\begin{figure*}
\centering
\includegraphics[width=.6\linewidth]{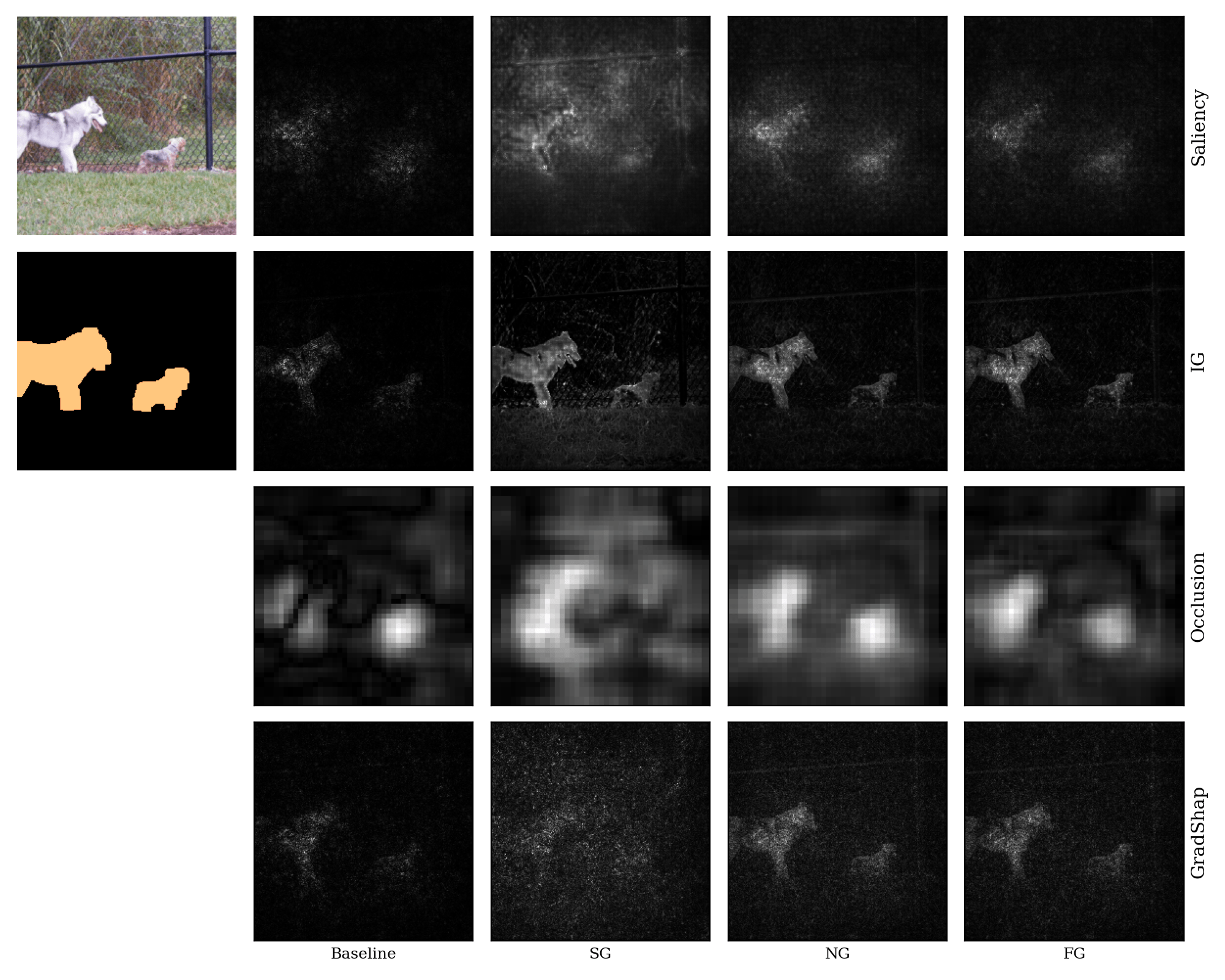}
\caption{Comparison of different enhancing methods for class "dog" for ImageNet dataset, using predictions from a pre-trained ResNet-18 classifier. We can observe that NG and FG methods are better in attributing relevance to the right dog, in comparison to the Baseline and SG methods.}
  \label{fig:BayesianLearingConnection1}
\end{figure*}

\begin{figure*}
\centering
\includegraphics[width=.6\linewidth]{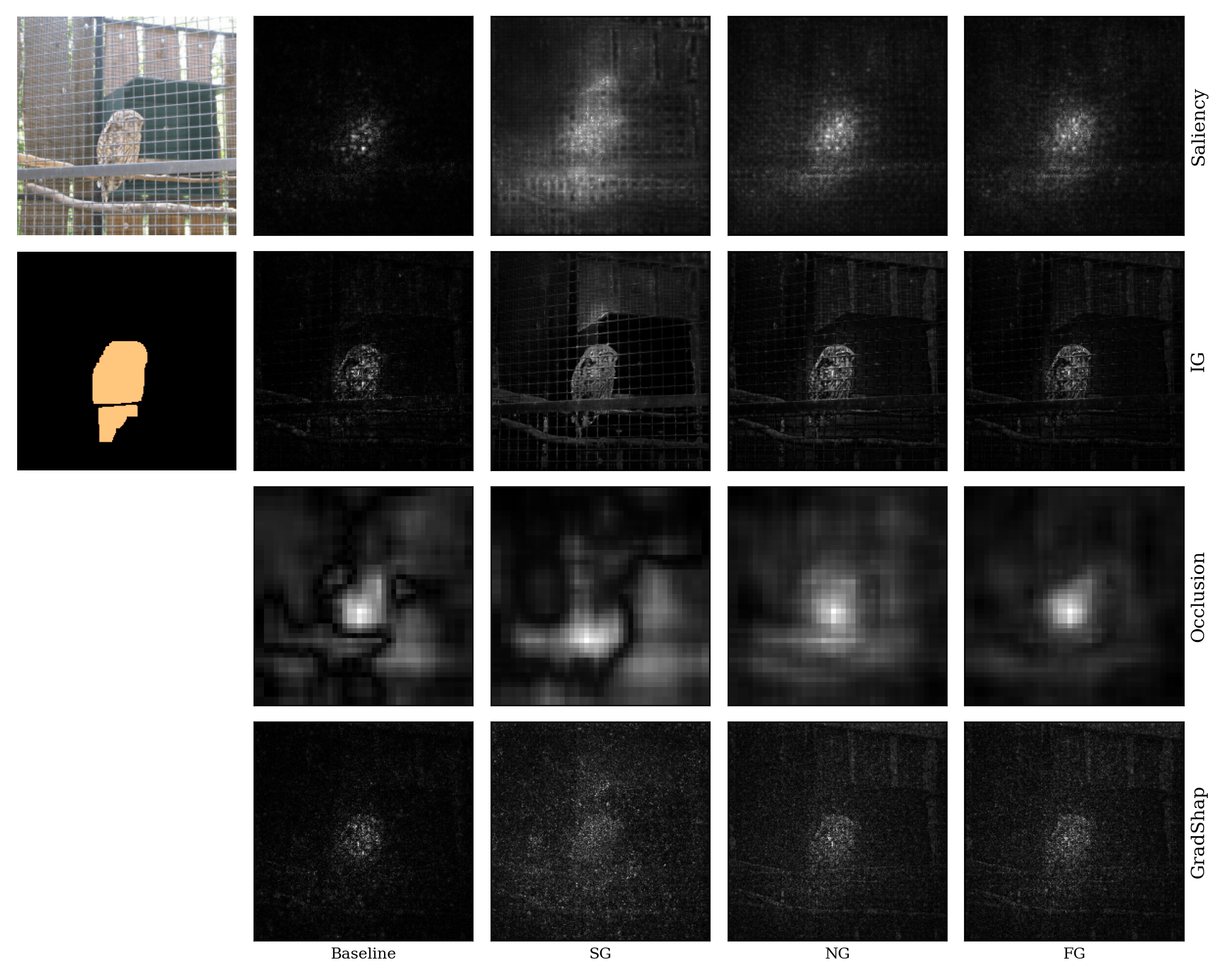}
\caption{Comparison of different enhancing methods for class "bird" for ImageNet dataset, using predictions from a pre-trained ResNet-18 classifier. We can observe that NG and FG place more attributional evidence at the object of interest, instead of the surroundings, like the Baseline and SG methods.}
\label{fig:BayesianLearingConnection2}
\end{figure*}

We provide further explanations of SmoothGrad, NoiseGrad, and FG for several Pascal VOC 2012 images and several different explanation methods in Figures \ref{fig:BayesianLearingConnection1}-\ref{fig:BayesianLearingConnection2}.

\subsubsection{Effect of number of optimization steps}

\begin{figure*}[!t]
\centering
\includegraphics[width=\linewidth]{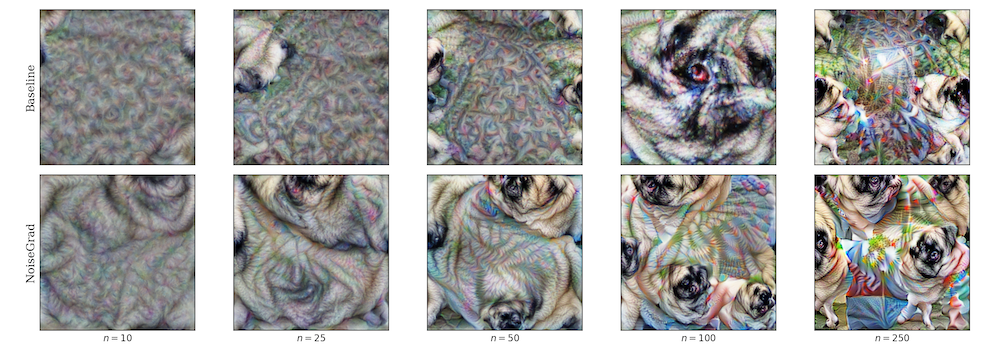}
\caption{Comparison of standard- (top row) and NoiseGrad-enhanced (bottom row) global explanations with respect to the different number of optimizations steps for class "Pug" from ImageNet dataset, predicted with Resnet-18 classifier.}
\label{fig:GlobalSteps1}
\end{figure*}

\begin{figure*}[!t]
\centering
\includegraphics[width=\linewidth]{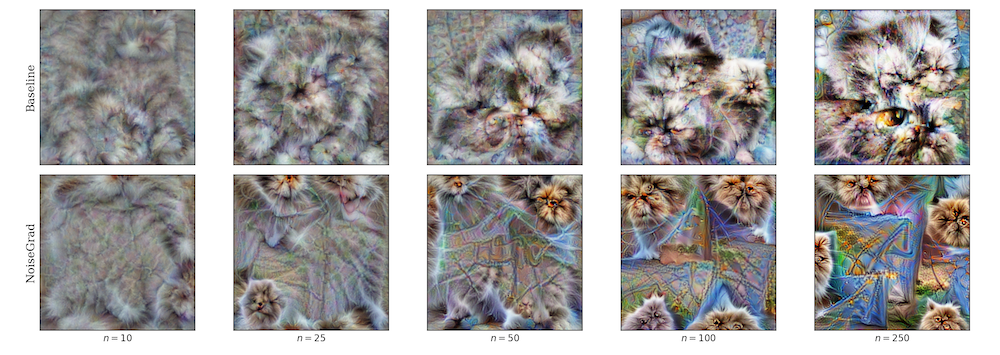}
\caption{Comparison of standard- (top row) and NoiseGrad-enhanced (bottom row) global explanations with respect to the different number of optimizations steps for class "Persian Cat" from ImageNet dataset, predicted with Resnet-18 classifier.}
\label{fig:GlobalSteps2}
\end{figure*}

One important hyper-parameter to any Activation Maximisation method is the number of optimization steps for the generation of the synthetic input. In Figures \ref{fig:GlobalSteps1}-\ref{fig:GlobalSteps2} we can observe the effect of the NoiseGrad enhancing procedure for different numbers of iterations for the optimization procedure. We can observe that enhancing effect is visible even for the low number of iterations.

\subsection{Computational cost}
Due to the nature of the proposed methods, the main cause for the computational complexity is the iterative perturbation of the models' weights.
Let us denote by $E$ the complexity of performing an explanation for a single input sample and a single model sample, and by $F$ the complexity of the drawing and adding/multiplying random noise to a single variable.  Then, the total complexities of SG, NG, and FG are $O(N (E + DF))$, $O(N (E + PF))$, and $O(N (E + (D + P)F))$, respectively, where $N$, $D$ and $P$ are the number of (input or model) samples, the input dimension, and the parameter dimension, respectively.
Therefore, the ratio between the computation time of NG (as well as FG) and that of SG is moderate for fast ($E > PF$) explanation methods (e.g., Saliency), and small for slow ($E < PF$) explanation methods (e.g., Integrated Gradient).
In our experiments, for the ResNet18 we have observed that NG is roughly 7.5 times slower, and 2  times slower with Saliency and Integrated Gradient methods respectfully. In both experiments, GPU was used and $N = 100$ for both methods, for CPU experiment computation time for both methods is approximately the same, as explanation complexity dominates the noise generation.
For costly ($E \gg PF$) explanation methods (e.g. LIME, Occlusion) computation time can be comparable.

\subsection{NoiseGrad and Global Explanations}

\begin{figure*}
\centering
\includegraphics[width=\linewidth]{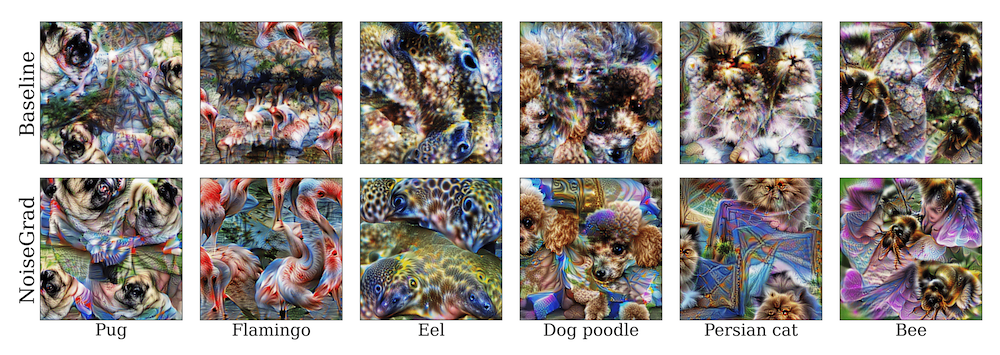}
\caption{Comparison of standard- (top row) and NoiseGrad-enhanced (bottom row) global explanations for different classes from ImageNet dataset. By comparing the Baseline and NoiseGrad visualizations, we can observe that the abstractions generated by NoiseGrad appears to be more semantically meaningful.}
\label{fig:GlobalAppendix1}
\end{figure*}

\begin{figure*}%
    \centering
    \subfloat[\centering Pug and Flamingo global explanations]{{\includegraphics[width=.35\linewidth]{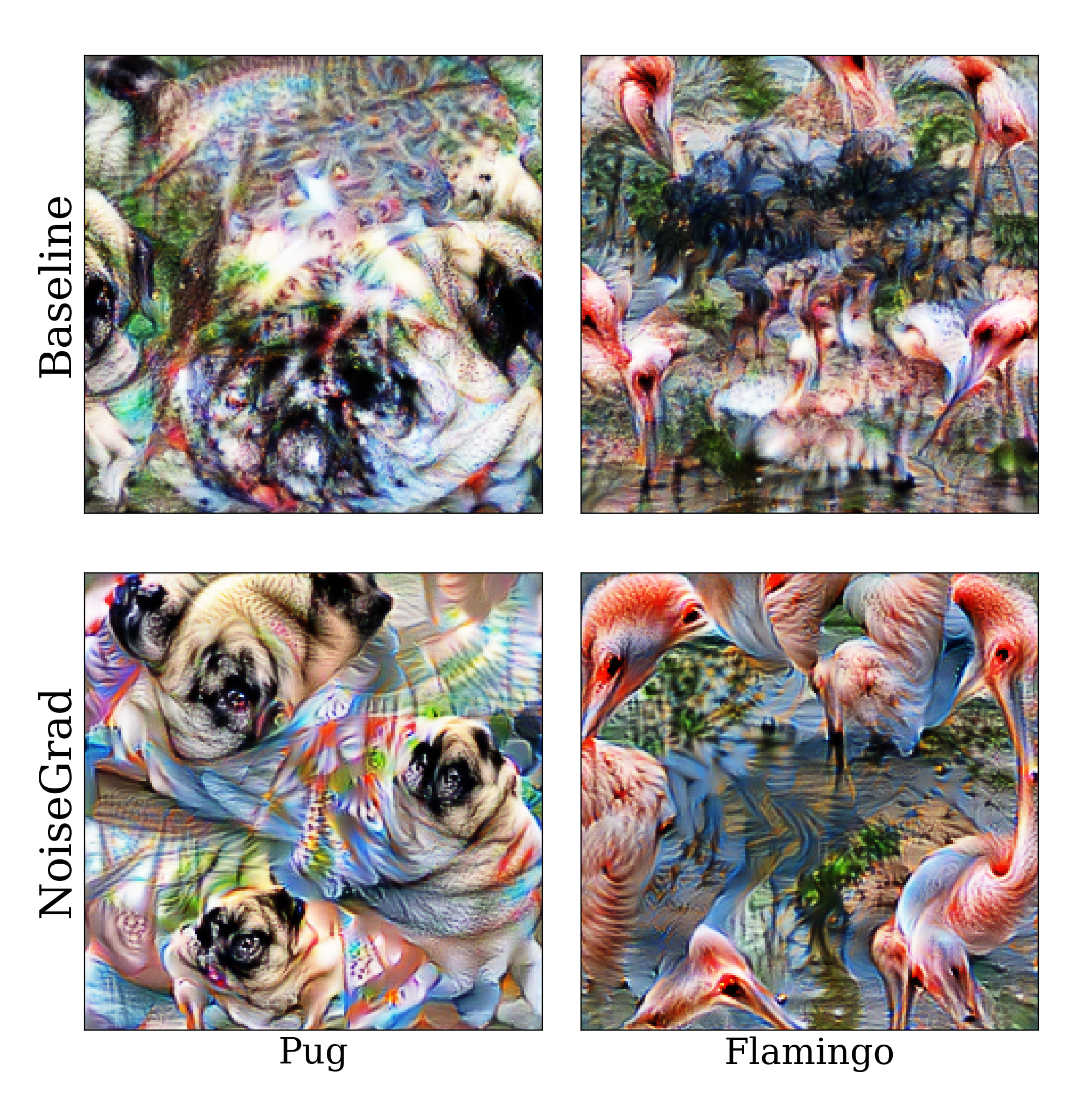} }}%
    \qquad
    \subfloat[\centering Pelican and Jellyfish global explanations 2]{{\includegraphics[width=.35\linewidth]{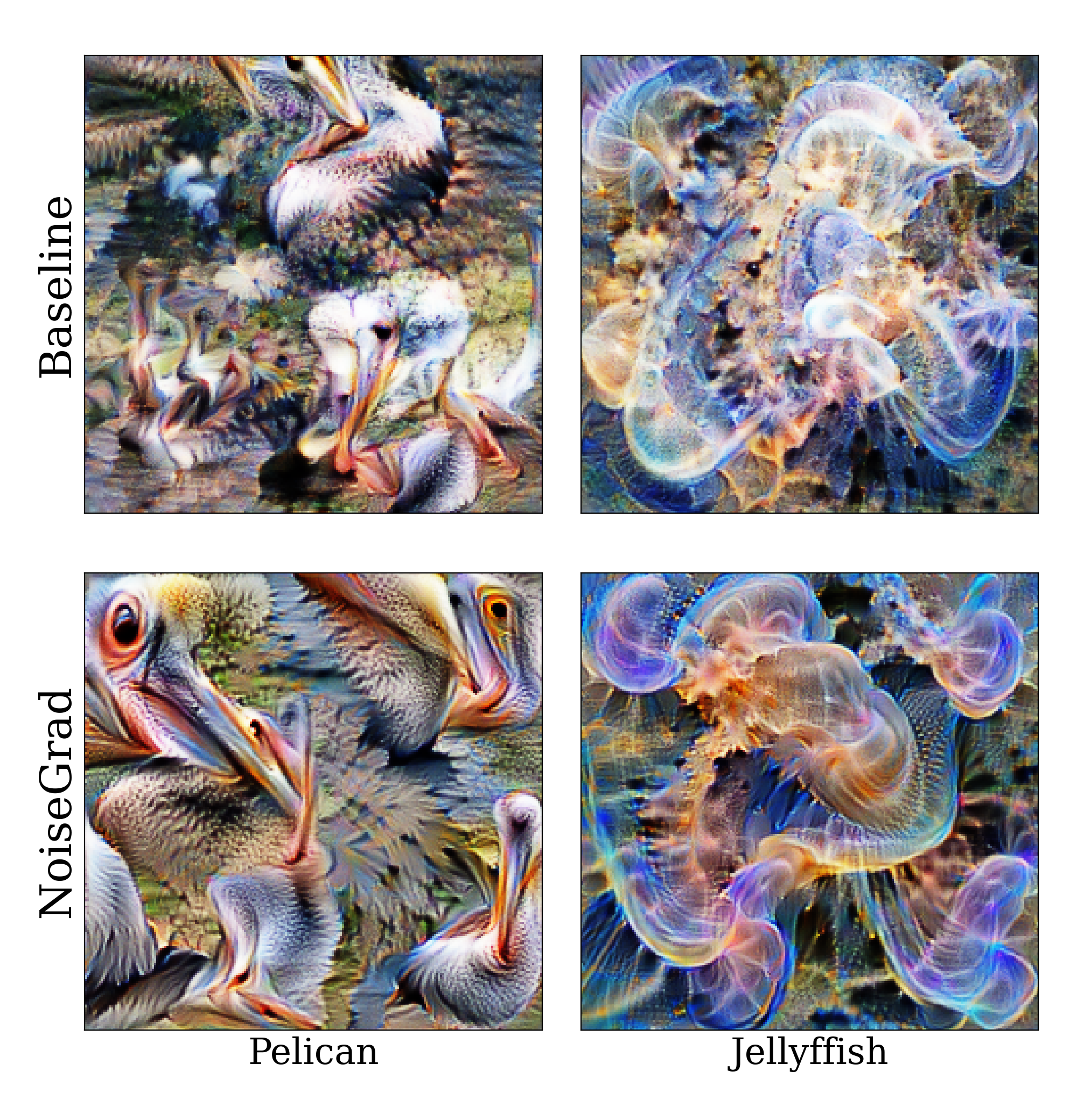} }}%
    \caption{Comparison of standard- (top row) and NoiseGrad-enhanced (bottom row) global explanations for different classes from ImageNet dataset. By comparing the Baseline and NoiseGrad visualizations, we can observe that the abstractions generated by NoiseGrad appears to be more semantically meaningful.}%
    \label{fig:GlobalAppendix2}%
\end{figure*}

We provide more examples of NoiseGrad's enhancing capabilities towards global explanations. We used a ResNet-18 \cite{he2015deep} pre-trained on Imagenet and we explained class logits with a Feature Visualisation \cite{olah2017feature} method.
Additional global explanations visualized in Figures \ref{fig:GlobalAppendix1}-\ref{fig:GlobalAppendix2}.

\end{document}